\documentclass[letterpaper, 10 pt, conference]{ieeeconf}  % Comment this line out if you need a4paper

\IEEEoverridecommandlockouts                          
\usepackage{hyperref}
\usepackage{amsmath}
\usepackage{graphicx}
\graphicspath{ {./images/} }
\usepackage{xcolor}

\usepackage{dblfloatfix}
\usepackage{caption}
\usepackage{subcaption}
\usepackage{algorithm}
\usepackage{algcompatible}
\usepackage{algpseudocode}
\usepackage{wrapfig}
\usepackage{multicol}
\usepackage{float}
\usepackage{tabularx}

\renewcommand{\baselinestretch}{.98}

\title{\LARGE \bf
GoalsEye: Learning High Speed Precision Table Tennis on a Physical Robot
}

\author{Tianli Ding$^{1,2}$, Laura Graesser$^{1,2}$, Saminda Abeyruwan$^{1}$, David B.\ D'Ambrosio$^{1}$,\\ Anish Shankar$^{1}$, Pierre Sermanet$^{1}$, Pannag R.\ Sanketi$^{1,3}$, Corey Lynch$^{1,2,3}$% <-this % stops a space
\thanks{$^{1}$Robotics at Google, Google Research, Mountain View, United States.}%
\thanks{$^{2}$Corresponding authors: tding@google.com, lauragraesser@google.com, coreylynch@google.com.}%
\thanks{$^{3}$Equal advising.}%
}

\begin{document}

\maketitle
\thispagestyle{empty}
\pagestyle{empty}

%%%%%%%%%%%%%%%%%%%%%%%%%%%%%%%%%%%%%%%%%%%%%%%%%%%%%%%%%%%%%%%%%%%%%%%%%%%%%%%%
\begin{abstract}

Learning goal conditioned control in the real world is a challenging open problem in robotics. Reinforcement learning systems have the potential to learn autonomously via trial-and-error, but in practice the costs of manual reward design, ensuring safe exploration, and hyperparameter tuning are often enough to preclude real world deployment. Imitation learning approaches, on the other hand, offer a simple way to learn control in the real world, but typically require costly curated demonstration data and lack a mechanism for continuous improvement. Recently, iterative imitation techniques have been shown to learn goal directed control from undirected demonstration data, and improve continuously via self-supervised goal reaching, but results thus far have been limited to simulated environments. In this work, we present evidence that iterative imitation learning can scale to goal-directed behavior on a real robot in a dynamic setting: high speed, precision table tennis (e.g. ``land the ball on this particular target"). We find that this approach offers a straightforward way to do continuous on-robot learning, without complexities such as reward design or sim-to-real transfer. It is also scalable---sample efficient enough to train on a physical robot in just a few hours. In real world evaluations, we find that the resulting policy can perform on par or better than amateur humans (with players sampled randomly from a robotics lab) at the task of returning the ball to specific targets on the table. Finally, we analyze the effect of an initial undirected bootstrap dataset size on performance, finding that a modest amount of unstructured demonstration data provided up-front drastically speeds up the convergence of a general purpose goal-reaching policy. See \url{https://sites.google.com/view/goals-eye} for videos.
\end{abstract}

%%%%%%%%%%%%%%%%%%%%%%%%%%%%%%%%%%%%%%%%%%%%%%%%%%%%%%%%%%%%%%%%%%%%%%%%%%%%%%%%
\section{Introduction}\label{sec:intro}
\begin{figure}[t]
\vspace{-2mm}
    \centering
    \begin{subfigure}{0.48\textwidth}
        \centering
        \smallskip
        \includegraphics[width=\textwidth]{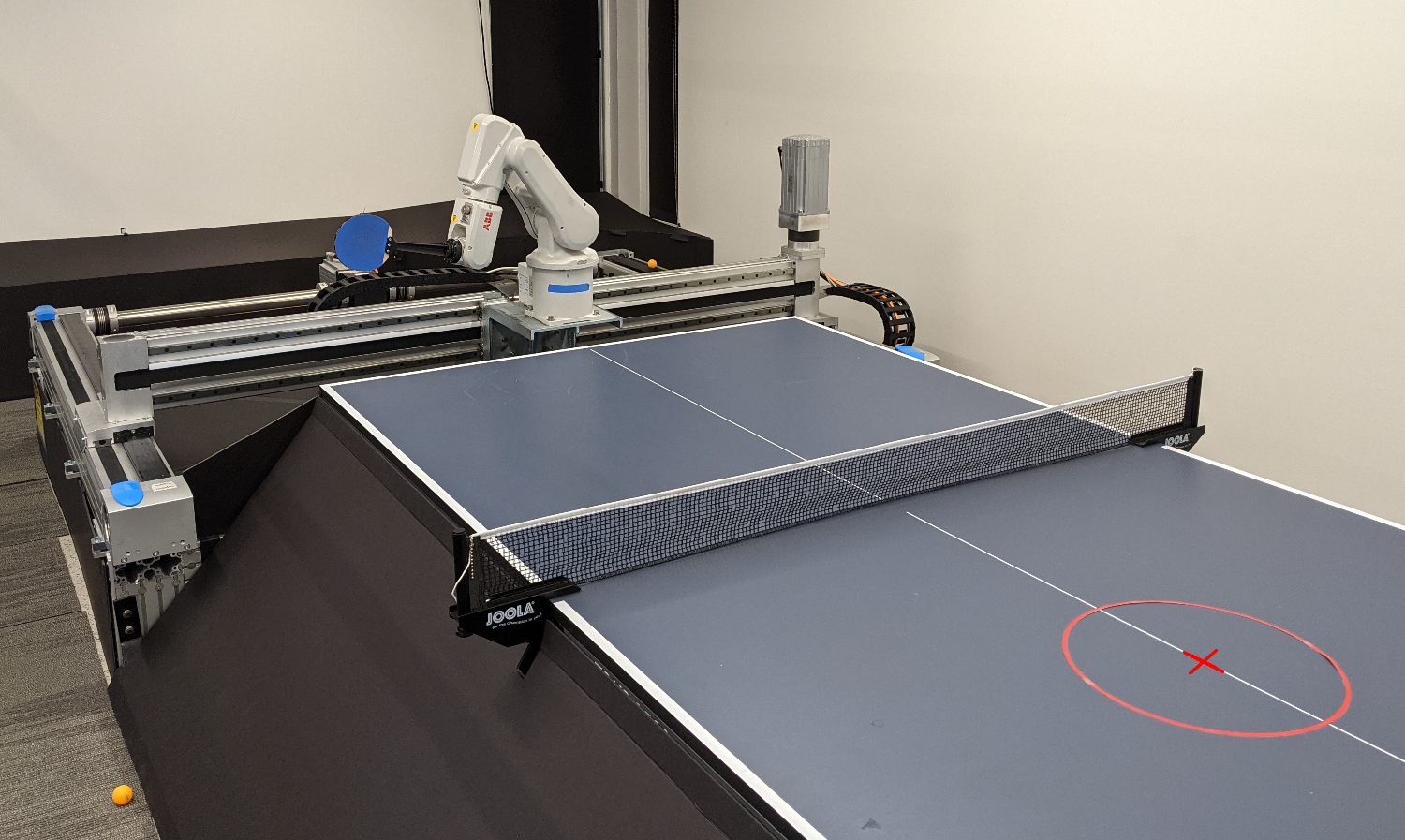}
        \caption{Example target location shown in red x, with an error margin defined by the circle (for display purposes only, no marks are present during training or evaluation).} 
        \label{fig:precision_ping_pong}
    \end{subfigure}%
    \hfill
    \begin{subfigure}[t]{0.24\textwidth}
        \centering
        \includegraphics[width=0.7\textwidth]{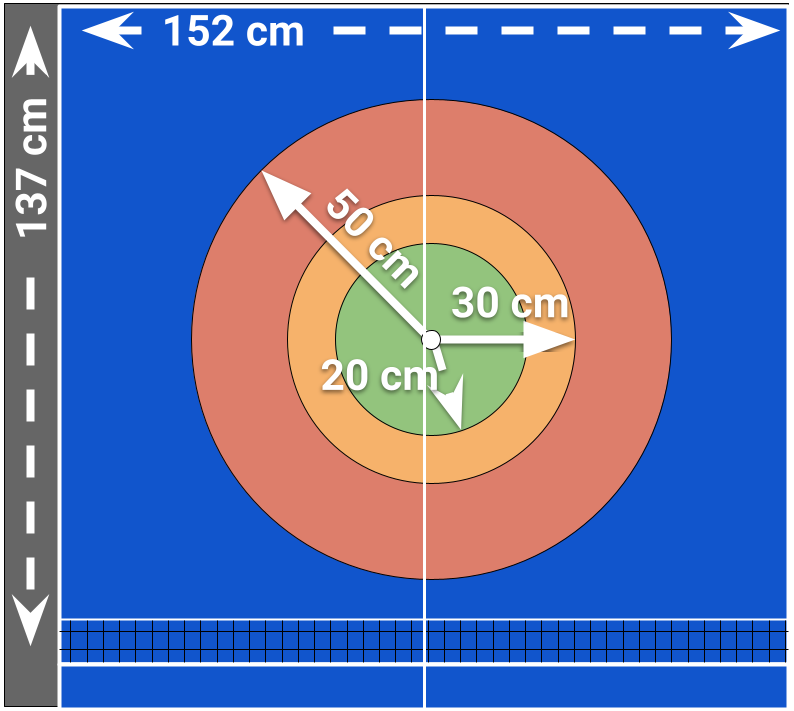}
        \caption{Goal thresholds}
    \label{fig:goalthresholds}
    \end{subfigure}
    \begin{subfigure}[t]{0.23\textwidth}
        \centering
        \includegraphics[width=\textwidth]{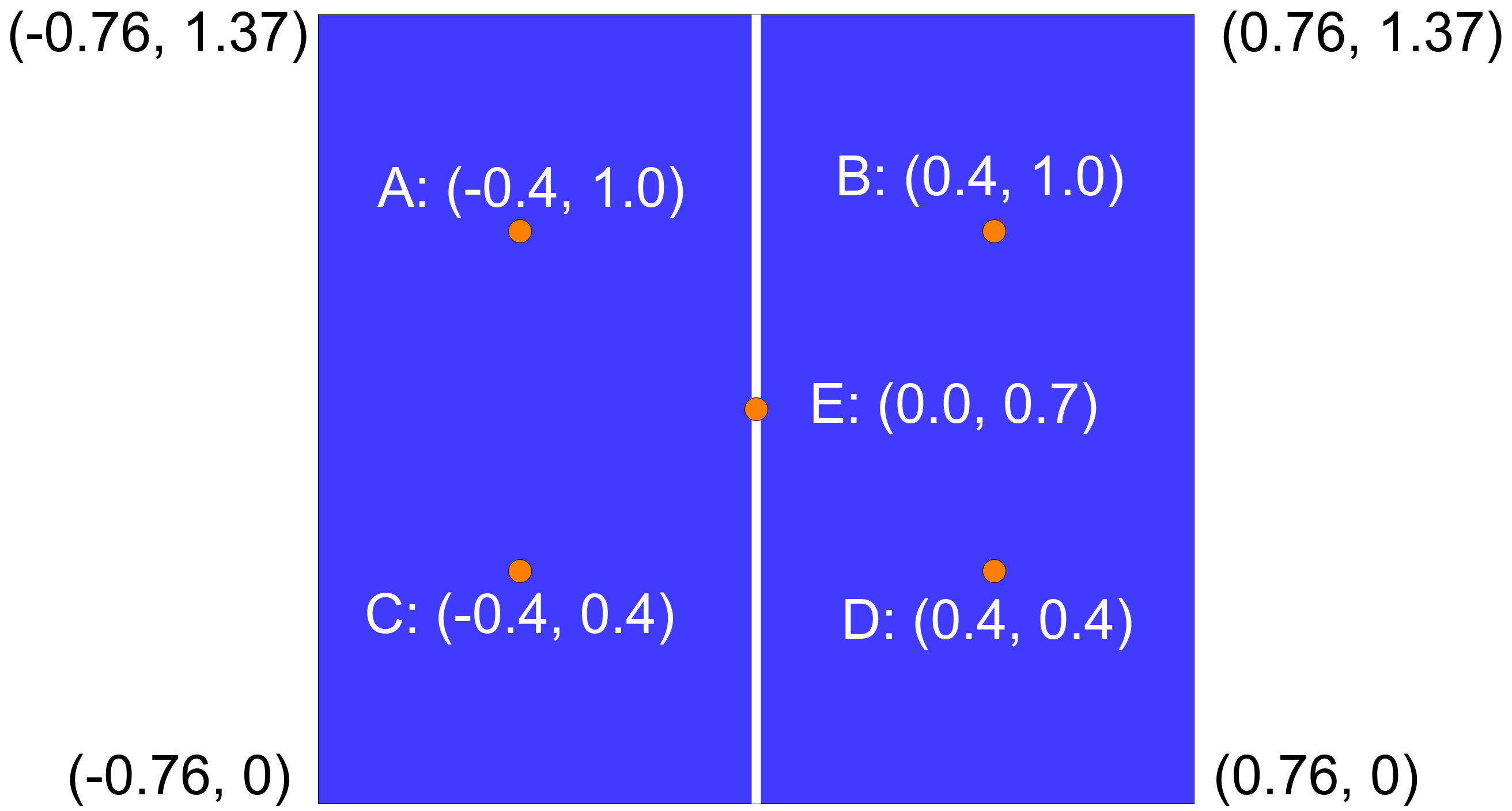}
        \caption{Five eval goal locations}
        \label{fig:fivegoals}
    \end{subfigure}
    \caption{Precision Table Tennis.  At test time, the robot must hit a ball fired at approximately $7~m / s$ to a commanded target location on the opposing side of the table.}
    \vspace{-6mm}
\end{figure}

Robot learning has been applied to a wide range of challenging real world tasks, including dexterous manipulation \cite{rubikscube, Mahler2019LearningAR}, legged locomotion \cite{Peng2020LearningAR, Tang2020LearningAL}, and grasping \cite{Kalashnikov2018QTOptSD, Xiao2020ThinkingWM}. It is less common, however, to see robotic learning applied to dynamic, high-acceleration, high-frequency tasks like \textit{precision table tennis} (Figure \ref{fig:precision_ping_pong}). Such settings put significant demands on a learning algorithm around safe exploration, accuracy, and sample efficiency. An outstanding question for robot learning is: can current techniques scale to meet the hard requirements of this setting? 

Consider the setup in Figure \ref{fig:precision_ping_pong}: a robot must issue 8-DOF continuous control commands in joint space at 20Hz to control an arm holding a paddle. The commanded behavior must precisely position and orient the paddle in time and space in order to connect with a ball fired at 7 meters per second. The right follow-through motion must be orchestrated in order to return the ball to the other side of the table. Strictly more difficult is the problem of learning to return the ball to an arbitrary \textit{target} location on the table, e.g. ``hit the back left corner" or ``land the ball just over the net on the right side".

Imitation Learning (IL) \cite{hussein2017imitation} provides a simple and stable approach to learning robot behavior, but requires access to demonstrations. Collecting expert demonstrations of precise goal targeting in such a high speed setting, say from teleoperation or kinesthetic teaching \cite{Muelling2012LearningSelectGen} is a complex engineering problem.  Attempting to learn precise table tennis by trial and error using reinforcement learning (RL) is a similarly difficult proposition given its sample inefficiency and that the random exploration that is typical at the beginning stages of RL may damage the robot. High-frequency control also results in long horizon episodes. These are among the biggest challenges facing current deep RL techniques \cite{Ibarz2021HowTT}. While many recent RL approaches successfully learn in simulation, then transfer to the real world \cite{Ho2020RetinaGANAO, rubikscube}, doing so in this setting remains difficult especially considering the requirement of precise, dynamic control. Here we restrict our focus to learning a hard dynamic problem directly on a physical robot without involving the complexities of sim-to-real transfer.

In this work, we consider what is the \textit{simplest} way to obtain goal conditioned control in a dynamic real world setting such as precision table tennis? Can one design effective alternatives to more intricate RL algorithms that perform well in this difficult setup? In pursuit of this question, we consider the necessity of different components in existing goal conditioned learning pipelines, both RL and IL. Surprisingly, we find that the synthesis of two existing techniques in iterative self-supervised imitation learning \cite{lfp}, \cite{ghosh2021learning} indeed scales to this setting. For ease of reference, we refer to this best performing approach throughout as \textit{GoalsEye}, a system for high-precision goal reaching table tennis, trained with goal conditioned behavior cloning plus self-supervised practice (GCBC$+$SSP).

We find that the essential ingredients of success are: \textbf{1) A minimal, but non-goal-directed ``bootstrap" dataset} to overcome an initial difficult exploration problem \cite{lfp}. \textbf{2) Relabeled goal conditioned imitation}: GoalsEye uses simple and sample efficient relabeled behavior cloning \cite{lfp, DBLP:journals/corr/abs-1906-05838, HER}, to train a goal-directed policy to reach any goal state in the dataset without reward information. \textbf{3) Iterative self-supervised goal reaching:} The agent improves continuously by giving itself random goals, then attempting to reach them using the current policy \cite{ghosh2021learning}. All attempts, including failures, are relabeled into a continuously expanding training set.

The main contributions of this work are: 
\textbf{1)} We introduce a setting of high-acceleration \textit{goal directed} table tennis on a physical robot.
\textbf{2)} We present GoalsEye, an iterative imitation learning system that can improve continuously in the real world to the point where it can execute precise, dynamic goal reaching behavior at or above amateur human performance. Our final system is able to control a physical robot at 20Hz to land 40\% of balls to within 20 centimeters of commanded targets at 6.5 m/s (see \url{https://sites.google.com/view/goals-eye} for videos).
\textbf{3)} We perform a large empirical study, both in simulation and in the real world, to determine what are the important components of success in this setting. We note that even though we present experimental results in the domain of robotic table tennis, nothing in our recipe is specific to table tennis and can be applied in principle to any task where a goal state can be specified at test time.

\section{Related Work}\label{sec:related}
\textbf{Robotic table tennis}. 
Table tennis has long served as as a particularly difficult benchmark for robotics. Research in robotic table tennis began in 1983 with a competition that had simplified rules and a smaller table \cite{Billingsley83}. This competition ran from 1983 to 1993 and several systems were developed \cite{Knight1986PingpongplayingRC,Hartley87,Hashimoto1987DevelopmentOP}; see \cite{Muelling2010Biomem} for a summary of these approaches. This problem remains far from solved.

Most approaches are model-based in that they explicitly model the ball and robot dynamics. The Omron Forpheus robot \cite{omron} is the current exemplar, achieving impressive results. These methods typically consist of several steps: identifying virtual hitting points from ball trajectories \cite{Miyazaki2002RealizationOT, Miyazaki2006LearningTD, Anderson1988ARP, Muelling2010SimulatingHT, Zhu2018TowardsHL, Huang2015LearningOS, Sun2011BalanceMG, Mahjourian2018HierarchicalPD}, predicting ball velocities by learning from data \cite{Miyazaki2002RealizationOT, Matsushima2003LearningTT, Matsushima2005ALA, Miyazaki2006LearningTD} or through a parameterized dynamics models \cite{Muelling2010Biomem, Muelling2010SimulatingHT, Zhu2018TowardsHL} calculating target paddle orientations and velocities, and finally generating robot trajectories leading to desired paddle targets \cite{Muelling2010Biomem, Muelling2010LearningTTMOMP, Muelling2012LearningSelectGen, Huang2016JointlyLT, Ko2018OnlineOT, Miyazaki2002RealizationOT, Matsushima2003LearningTT, Matsushima2005ALA, Miyazaki2006LearningTD, Muelling2010Biomem, Muelling2010LearningTTMOMP, Tebbe2018ATT, Gao2019MarkerlessRP}.

A number of methods do not model the robot dynamics explicitly. These approaches fall into two broad groups, those that utilize expert demonstrations \cite{Muelling2010LearningTTMOMP, Muelling2012LearningSelectGen, Huang2016JointlyLT, Akrour2016ModelFreeTP, LFSD-GT} and those that do not \cite{GaoPPOES2020, Zhu2018TowardsHL, buchler, SERL_tebbe}. Like our best performing method, \cite{LFSD-GT} is capable of learning from sub-optimal demonstrations. However, the approach has no mechanism to continuously improve beyond the demonstration data. In \cite{Muelling2012LearningSelectGen}, authors demonstrate a system that learns cooperative table tennis by creating a library of primitive motions using kinesthetic teaching to constrain learning. In a similar spirit, we collect an initial dataset of non-goal-directed demonstration data of how to make contact and return the ball to bootstrap autonomous learning.

Reinforcement learning (RL) is a common approach for table tennis methods that do not utilize demonstrations. Methods range from framing the problem as a single-step bandit \cite{Zhu2018TowardsHL} to temporarily extended policies controlling the robot in joint space \cite{GaoPPOES2020} using on-policy RL, to Hierarchical RL (HRL) \cite{PolicySearchHRL}. Of particular interest is \cite{buchler}, which utilizes muscular soft robots to facilitate safe exploration and learn RL policies from scratch on a real robot.

\textbf{Goal conditioned imitation learning}. 
While many of the above methods have been shown to scale to \textit{undirected} table tennis, few have tackled the problem of \textit{goal-directed} table tennis. Goal directed control is an active area of robot learning, with many recent examples in both IL and RL \cite{nair2018visual,lfp,DBLP:journals/corr/abs-1906-05838,actionable_models}. Given the complexities of even single task real world robot learning \cite{zhu2020ingredients} finding simple methods that scale to goal-directed real world behavior remains an open question. While goal-conditioned imitation learning \cite{lfp, DBLP:journals/corr/abs-1906-05838} offers a simple approach to multitask control, no instances yet have been shown to scale to hard physical problems like the one studied in this work, being largely validated in simulation instead. We find surprisingly that the simple combination of two existing IL methods \cite{lfp}, \cite{ghosh2021learning} indeed scales to this setting, while being able to 1) learn from less burdensome suboptimal (in the sense of being non goal-directed) demonstrations, 2) use relabeled learning to learn goal-reaching without rewards, and 3) continuously self-improve beyond the initial data by using self-supervised goal reaching.

\textbf{Empirical studies in scaling robot learning}. 
Like many works in robot learning \cite{fujimoto2019benchmarking,dulac2020empirical,agarwal2019striving}, ours studies empirically whether existing methods scale to new and harder robotic problems than the ones originally studied. For example, studies such as \cite{haarnoja2018soft} found new evidence that existing algorithms (e.g SAC), previously only studied in simulation, indeed scaled to hard problems such as real world quadrupedal locomotion. Similarly, recent empirical studies have shown that well motivated prior ideas did \textit{not} scale to more difficult robotic setups \cite{andrychowicz2020matters,yu2020meta}. For example, the recent work \cite{mandlekar2021matters} reported the surprising finding that simply switching from RL-agent generated offline data to human-collected offline data caused most offline RL approaches to degrade substantially. Surprising empirical phenomena such as this motivate studies like ours which help assess if the claims of existing methods generalize beyond the setups for which the original papers were written.

\section{Method}\label{sec:method}
	
The approach we study in this work consists of three elements: 1) a non-goal-directed ``bootstrap" dataset, 2) goal conditioned imitation with sample relabeling, and 3) continuous improvement through iterative self-supervised goal reaching. An overview of the method is given in Algorithm \ref{alg:main} and we now discuss each of the three elements in turn.

\begin{algorithm}
\caption{GoalsEye Algorithm}\label{alg:main}
\begin{algorithmic}
\State $StepsBetweenSSP > 0$
\State $NumSspPerIter > 0$
\State Initialize $Cache$
\State Initialize $Policy$
\State $Cache \gets Initial Demos$
\State $Step \gets 0$
\While {$True$}
    \State Sample $Batch$ from $Cache$
    \State Train $Policy$ using $Batch$
    \State $Step \gets Step + 1$
    \If{$Step \mod StepsBetweenSSP = 0$ }
        \State $NumRollouts \gets 0$
        \While {$NumRollouts \leq NumSspPerIter$}
            \State Sample random $Goal$
            \State Rollout current $Policy$, try to reach $Goal$
            \State Relabel $Goal$ with actual ball landing $X, Y$
            \State Write episode to $Cache$
            \State $NumRollouts \gets NumRollouts + 1$
        \EndWhile
    \EndIf
\EndWhile
\end{algorithmic}
\end{algorithm}

\textbf{Bootstrapping from non-goal-directed data}. We assume that policies have access to a small number of demonstrations with the following \emph{minimum} properties:

1) The demonstrations need to be skillful enough to overcome an initial hard exploration problem but they do not need to be optimal. For goal-directed robotic table tennis we require a demonstration to hit the ball it is thrown, and to successfully return the ball to the opponent's side of the table with a non negligible probability. It is not necessary to be able to return the ball to a goal with any amount of accuracy.

\begin{figure*}[!tp]
    %  \vspace{-2mm}
     \centering
     \begin{subfigure}[t]{0.3\textwidth}
         \centering
         \includegraphics[width=\textwidth]{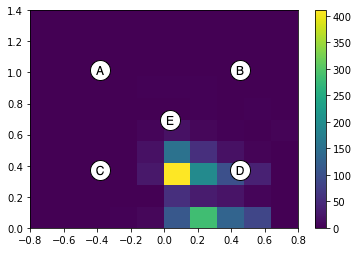}
         \caption{The initial "narrow" policy.}
         \label{fig:datanarrow}
     \end{subfigure}%
     \hfill
     \begin{subfigure}[t]{0.3\textwidth}
         \centering
         \includegraphics[width=\textwidth]{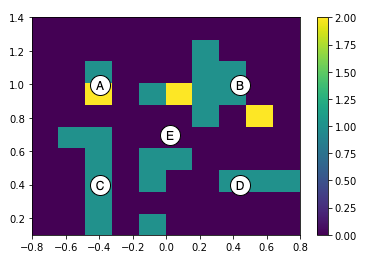}
         \caption{The sparse, bootstrap dataset.}
         \label{fig:databootstrap}
     \end{subfigure}
     \hfill
     \begin{subfigure}[t]{0.3\textwidth}
         \centering
         \includegraphics[width=\textwidth]{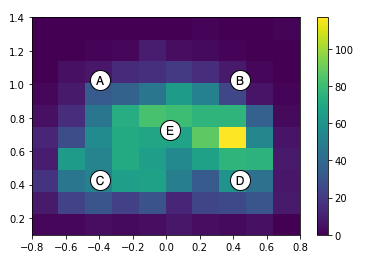}
         \caption{Final policy.}
         \label{fig:datafinal}
     \end{subfigure}
     \caption{Comparing the ball landing distribution of the initial "narrow" policy, the bootstrap data, and the final policy after training on the physical robotic system.}
     \label{fig:bootstrap_dataset}
    \vspace{-4mm}
\end{figure*}

2) Qualitatively, a dataset of initial demonstrations which includes more varied state-action trajectories can result in easier training processes with better results. In our problem, we sought to create a dataset in which the ball landing locations spanned the opponents side of the table (see Figure \ref{fig:databootstrap}). Note that this second requirement can be bootstrapped from a narrow set of initial demonstrations if varied demonstrations are difficult to obtain.

\textbf{Relabeled imitation learning}. The training dataset consists of a set of non-goal-directed trajectories $T$ where $s$ is an observation describing the state of the environment and $a$ is a robot command: $T = (t_1, t_2, ..., t_N)$ and $t_N = (s_{N1}, a_{N1}, s_{N2}, a_{N2}..., s_{Nn}))$.

We apply hindsight relabeling \cite{lfp} to the non-goal-directed demonstrations to transform them into goal-conditioned demonstrations by assuming that the final state in the trajectory was the goal the policy was actually trying to reach. Now we have $T = ((t_1, g_1=s_{1n}), (t_2, , g_2=s_{2n}) ..., (t_N, g_N=s_{Nn}))$. Finally, the dataset is augmented each training step by randomly sampling a sub-sequence of length $k$. In simulation we set $k=96$, and in the real world we set $k=16$. In both we ensure that the timestamp of the ball hit point is included in the sampled sub-sequence.

A parameterized policy $\pi_{\theta}(s, g)$ is trained to imitate $T$ by minimizing the mean squared error between the commanded and realized joint positions per step, given an observation. That is;
\vspace{-2em}

\begin{equation}
\begin{aligned}
\min\limits_{\theta} \frac{1}{Nn} \sum_{i=1}^N \sum_{j=1}^n (a_{ij} - \pi_{\theta}(s_{ij}, g_{i}))^2
\end{aligned}
\end{equation}

For the problem of robot table tennis, the location of desired ball landing point is sufficient to describe the goal state. Demonstrations correspond to a single ball throw and return. Whilst the objective during evaluation is to return each ball to a specific goal chosen randomly on the opponent's side of the table, we found that extending the permissible goals in training beyond the physical boundaries of the table improved performance in simulation by ~20\% when the goal is less than 20cm from the edges.

\textbf{Continuous improvement through self-supervised practice}. Whilst a policy is training, it continuously self-practices in the environment, by attempting to reach goals sampled random uniformly from the opponent's side of the table. Each practice episode is processed and filtered out if it is not good enough (i.e. if the ball was not hit). If it was "good" it is relabeled and added to the dataset from which the policy is training. This process facilitates continuous improvement by expanding the training dataset over time leading to more precise goal-reaching.

\subsection{Building up the initial bootstrap dataset}

Generating a demonstration dataset with the desired properties can be difficult, expensive, or time consuming. Unlike lower control frequency real world tasks \cite{zhang2018deep}, obtaining quality teleoperation data for high acceleration tasks such as precision table tennis is itself a difficult engineering challenge. Previous approaches to generating demonstration data in the table tennis setting \cite{buchler} involve customized hybrid sim-to-real training, including a ``rebound model", whose parameters needed to be tuned empirically to enable accurate sim2real transfer. 

Surprisingly, counter to what has been hypothesized in prior work \cite{buchler}, we found that it is possible to train a standard ES \cite{NES, Nesterov2017RandomGM, salimans2017evolution, ARS, ChoromanskiRSTW18} policy to convergence fully in simulation, then apply it directly to the real robot as a means of obtaining an initial dataset. We found empirically that this approach was sufficient to get a real robot to make safe contact with the ball, bootstrapping further autonomous learning. However, we note that the initial policy obtained by ES was narrow, causing most of the examples to land close to the net in the right half of the opponent's side of the table (see Figure \ref{fig:datanarrow}). To overcome this limitation and more effectively cover the test goal distribution, we applied steps (2) relabeled imitation learning and (3) self-practice, setting the goal during policy rollouts higher up the table on the opponent side, with the intention of shifting the distribution up. Then we sought to expand the goal coverage by changing the goal distribution to the entirety of the opponent's side of the table. Additionally we perturbed 4 of the 8 robot joint angles during the data gathering stage in order to increase the variety of the ball landing points. The final demonstration dataset ball landing distribution is shown in Figure \ref{fig:databootstrap}.
\section{System Description}\label{sec:system}

\textbf{Hardware}. \textit{Player Robot:} The player robot (Figure \ref{fig:precision_ping_pong}) is a combination of an ABB IRB 120T 6-DOF robotic arm mounted to a two-dimensional Festo linear actuator, creating an 8-DOF system.  The robot arm's end effector is a standard table tennis paddle with the handle removed attached to a 174.26mm extension.  The arm is controlled with ABB's External Guided Motion (EGM) interface at approximately 240Hz by specifying joint and velocity targets \cite{abb2020egm}. The 2D actuator is independently controlled at up to 100Hz with position target commands for each axis at a fixed velocity through Festo's custom Modbus interface.  Feedback from the robots is received at the command rate.  When no ball is in play, the arm of the robot returns to a home position and the linear actuator remains fixed, otherwise the arm and actuator are free to move as defined by the learned policy.

\textit{Vision System:} The ball location is determined through a stereo pair of Ximea MQ013CG-ON cameras running at 125Hz via a recurrent 2D detector model trained on $\approx2$ hours of ball video data and 3D tracker.

The ball position and robot feedback are interpolated to the 20Hz the policy inferences on.

\textbf{Simulation Studies}. The physical system is modelled in simulation using PyBullet \cite{coumans2019}. The simulation uses a simplified ball dynamics model that includes drag but excludes spin. The ball throws are generated by randomly sampling an initial ball thrower position from the opponent side, aimed at a ball landing position on the player robot side of the table. Then the full initial velocity vector is solved and throws a ball approximately  $7$~m/s towards the robot side of the table. At the start of each ball throw (start of an episode), the arm is initialized to a central pose. The initial pose is perturbed to prevent overfitting.

\textbf{Policy architecture}. The GoalsEye policy is a 2-layer LSTM with a single fully-connected output layer. The size of each hidden layer is 1024, the output layer is 8, corresponding to the 8 robot joints. An observation consists of 16 elements in simulation; the ball xyz position (3) and velocity (3), robot joint positions (8), and the goal (2). On the real system an observation consists of 13 elements because ball velocity is not available. Policies control the robot at 100Hz in simulation and at 20Hz on the real robot. The maximum sequence length is 96 timesteps in simulation and 16 on the real robot (almost a full episode).
\section{Results}
\label{sec:result}

We present an overview of our results across all tasks, settings, and methods in Table \ref{table:results}. First we evaluate our approach on a challenging task in simulation: \textbf{any-ball goal-reaching} (Table \ref{tab:result_overview_sim}). Given any ball throw\footnote{Modelled by sampling ball throws from a wide distribution of initial positions and velocities.} return it to any location on the opponent's table side with high precision. That means that goals are sampled from the entire opponent's side of the table. Then we train a policy on a physical robot for a simplified version of this task: \textbf{narrow-ball goal-reaching} (Table \ref{tab:result_overview_real}). Given a forehand ball throw with the same launch position and a narrow range of velocities, return the ball to any location on the opponent's side of the table.

\textbf{Evaluation metrics}. We evaluate policies by calculating the percentage of balls that land within 30cm, or 20cm of the goal. These thresholds correspond to an area covering 14\% and 6\% respectively of the total goal area (see Figure \ref{fig:goalthresholds}). Each method was trained in simulation using 5 separately seeded runs, with each training run evaluated with 200 randomly sampled goals per checkpoint. We trained a single final policy on the physical robot. Additionally we compare the physical robot performance with human amateurs by setting five specific goals (see Figure \ref{fig:fivegoals}) for both humans and the robot to reach.

\begin{table}[t!]
\vspace{2mm}
    \caption{Summary of all methods on simulated and real environments.}
    \label{table:results}
    \begin{subtable}{0.95\linewidth}
      \centering
        \begin{tabular}{|l|c|c|c|}
        \hline
        Method & Dist to Goal (m) & $\le$30cm (\%) & $\le$20cm (\%) \\
        \hline
        GoalsEye &     \textbf{0.84 +/- 0.08}  & \textbf{21 +/- 2}   & \textbf{11 +/- 2} \\
        LFP      &     1.03 +/- 0.05     & 09 +/- 1   & 04 +/- 0.5 \\
        GCSL     &     nan           & 00 +/- 0        & 00 +/- 0 \\
        PPO     &     1.47 +/- 0.22     & 04 +/- 3   & 01 +/- 1 \\
        ES       &     nan           & 00 +/- 0        & 00 +/- 0 \\
        SAC       &     nan           & 00 +/- 0        & 00 +/- 0 \\
        QT-OPT   &     nan           & 00 +/- 0        & 00 +/- 0 \\
        \hline
        \end{tabular}
        \caption{Simulated any ball, all goal task.}
        \label{tab:result_overview_sim}
    \end{subtable}
    \begin{subtable}{.95\linewidth}
      \centering
        \begin{tabular}{|l|c|c|}
        \hline
        Method & $\le$30cm (\%) & $\le$20cm (\%) \\
        \hline
        GoalsEye  &     \textbf{61}              &       \textbf{41}         \\
        LFP       &     56         &       34          \\
        Human Avg.   &     33              &       14          \\
        \hline
        \end{tabular}
        \caption{Real world narrow ball, 5 goal task.}
        \label{tab:result_overview_real}
    \end{subtable}
    \vspace{-6mm}
\end{table}

\begin{figure}[!t]
\vspace{2mm}
     \centering
     \begin{subfigure}[b]{0.23\textwidth}
         \centering
         \includegraphics[width=\textwidth]{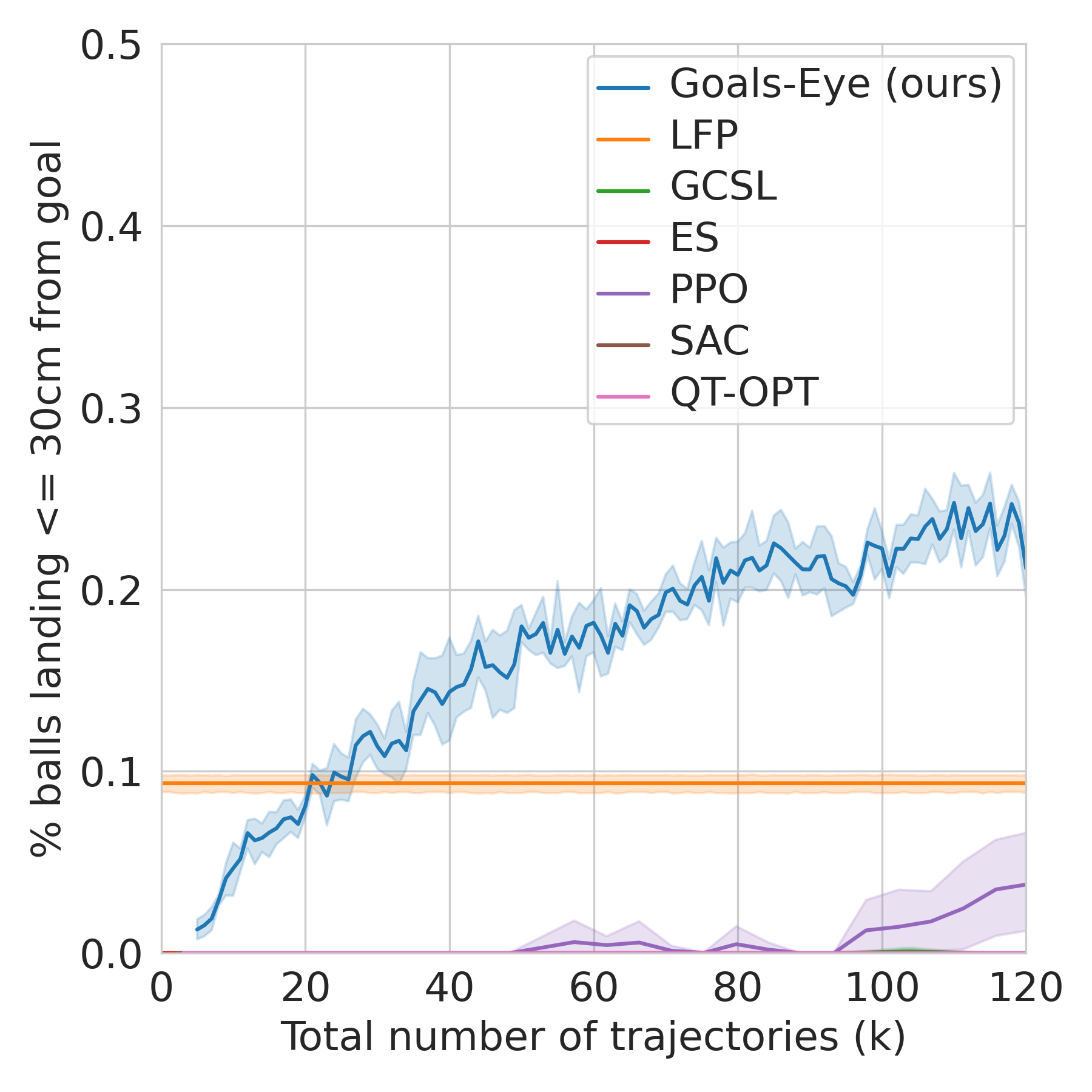}
         \caption{$\le$30cm to goal \newline vs. trajectories}
         \label{fig:sim_main30_120k}
     \end{subfigure}%
     \hfill
     \begin{subfigure}[b]{0.23\textwidth}
         \centering
         \includegraphics[width=\textwidth]{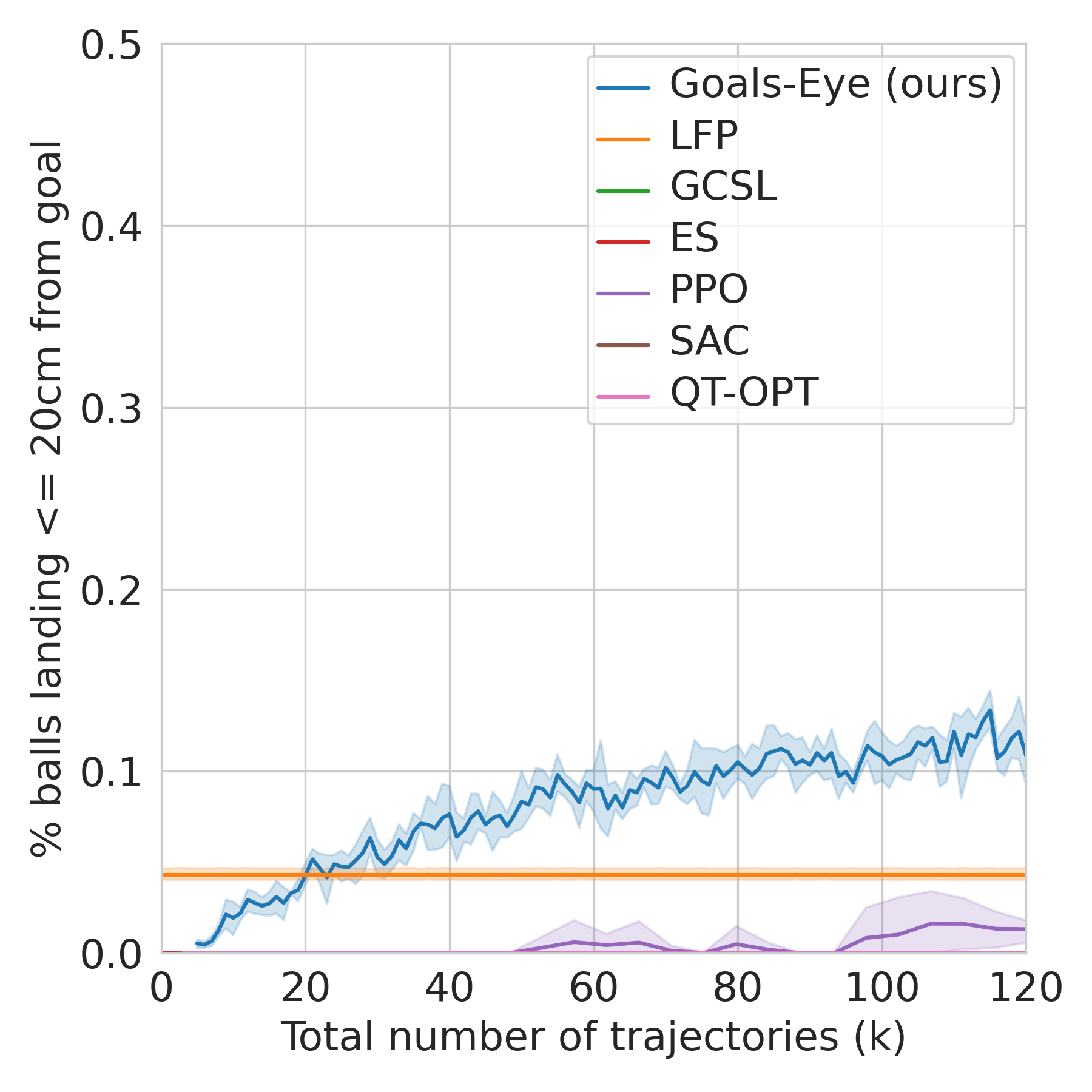}
         \caption{$\le$20cm to goal \newline vs. trajectories}
         \label{fig:sim_main20_120k}
     \end{subfigure}%
     \hfill
     \begin{subfigure}[b]{0.23\textwidth}
         \centering
         \includegraphics[width=\textwidth]{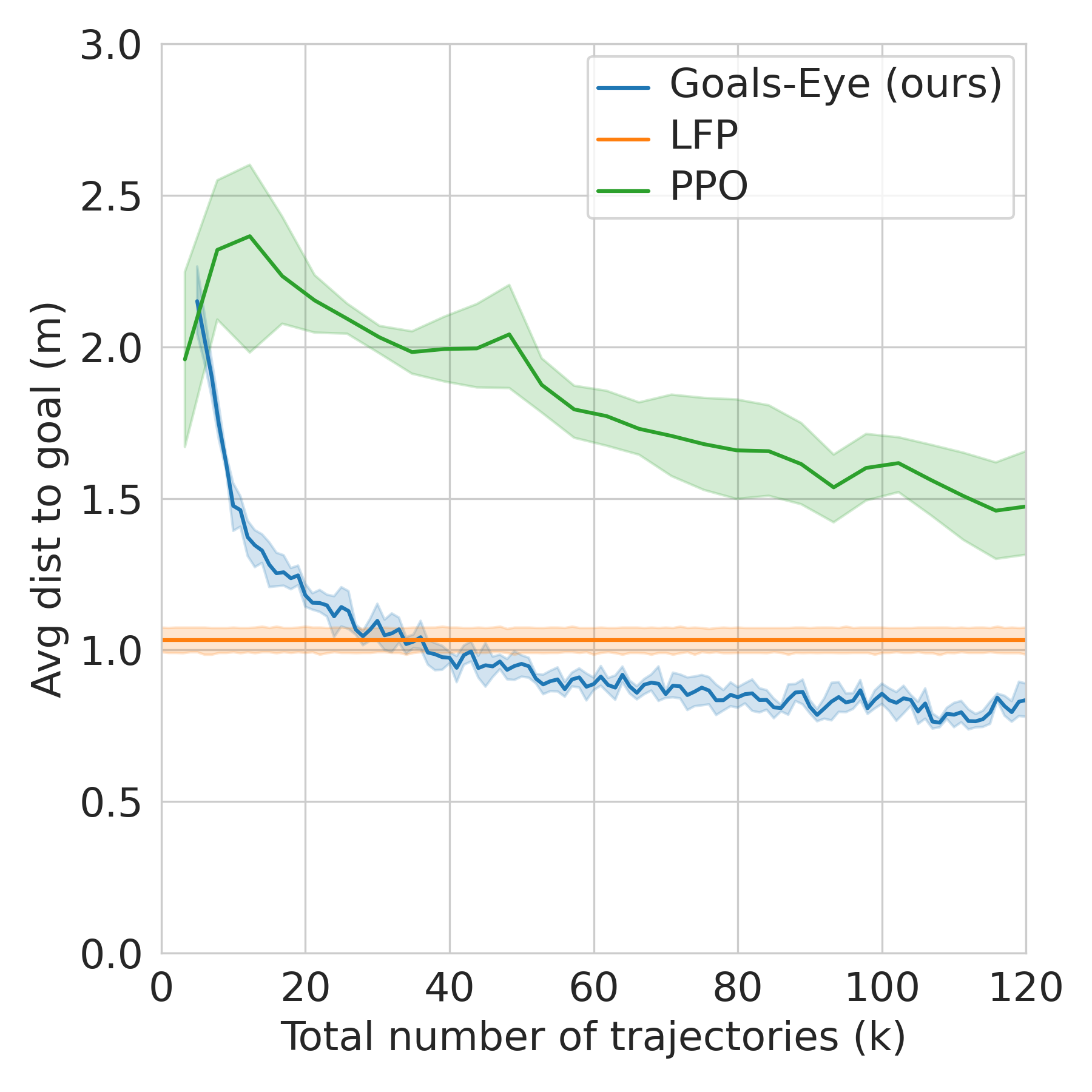}
         \caption{Distance to goal (m) \newline vs. trajectories}
         \label{fig:sim_main_dist_120k}
     \end{subfigure}
\caption{Comparison of all methods in simulation on the any ball goal-reaching task.}
\label{fig:main_sim}
\vspace{-6mm}
\end{figure}

\textbf{Any ball goal-reaching in simulation}. Table \ref{tab:result_overview_sim} presents the mean and standard deviation of 5 seeds for each of the 7 algorithms considered after 120k trajectories (equivalent to $\approx$ 6 days of continuous training on a physical robot), except for LFP \cite{lfp} which by definition only uses the initial demonstrations. Figure \ref{fig:main_sim} shows the corresponding learning curves. On this task GoalsEye achieves 21\% goal-reaching success within 30cm of the target and 11\% within 20cm of the target after 120k total trajectories. 3.6k of these trajectories were the initial demonstrations, and the remainder were generated through self-supervised practice. GoalsEye improves $>2\times$ on the 30cm metric and $\approx 3\times$ on the 20cm metric compared to LFP which lacks a mechanism for continuous improvement. On this task GCSL \cite{ghosh2021learning} fails to overcome the initial hard exploration problem as described in Section \ref{sec:method}. One reason why the exploration problem is particularly hard in this setting is because in simulation the robot is initialized with the front edge of the paddle facing forward, reducing the probability of hitting the ball through random exploration.

Given the prevalence of RL in robotic learning we also compare GoalsEye with two on-policy and two off-policy RL algorithms as a baseline on our simulated tasks. Note that the data efficiency of these methods and the requirement of initial random exploration excludes their application on our real world tasks. These policies are not trained with demonstrations. Instead they are trained with a reward function containing a number of elements; how close the ball lands to the target, whether the paddle makes contact with the ball, whether the agent lands the ball on the opponent's side of the table, and a number of rewards encouraging good style including arm pose and policy smoothness.

ES and PPO \cite{DBLP:journals/corr/SchulmanWDRK17} are two widely used on-policy RL algorithms. In the low data setting limited to 120k trajectories (Figure \ref{fig:main_sim}) ES fails to learn anything, whilst PPO just reaches 4\% success rate on the 30cm metric. However, given enough environment trajectories both algorithms achieve comparable or better performance compared with GoalsEye when trained from scratch without any demonstration data (see Figure \ref{app:main_sim_ext}). However they are significantly less sample efficient, requiring $\approx2\times$ (PPO) and $\approx150\times$ (ES) more trajectories in the environment to do so.

We also compare against SAC \cite{sac} and QT-OPT \cite{Kalashnikov2018QTOptSD}, two commonly used off-policy RL algorithms. Despite significant effort, we have so far not succeeded in training a successful policy using either method on this goal-reaching task (as shown by scores of 0 in Figure \ref{fig:main_sim} for both algorithms) or on the simpler task of returning a ball to the opponent's side of the table. We do not claim that off-policy methods do not work on this problem, however we do observe that it appears significantly more difficult to train policies using this approach compared with either on-policy RL or GoalsEye.

\begin{figure}[!t]
\vspace{2mm}
     \centering
     \begin{subfigure}[t]{0.235\textwidth}
         \centering
         \includegraphics[width=0.8\textwidth]{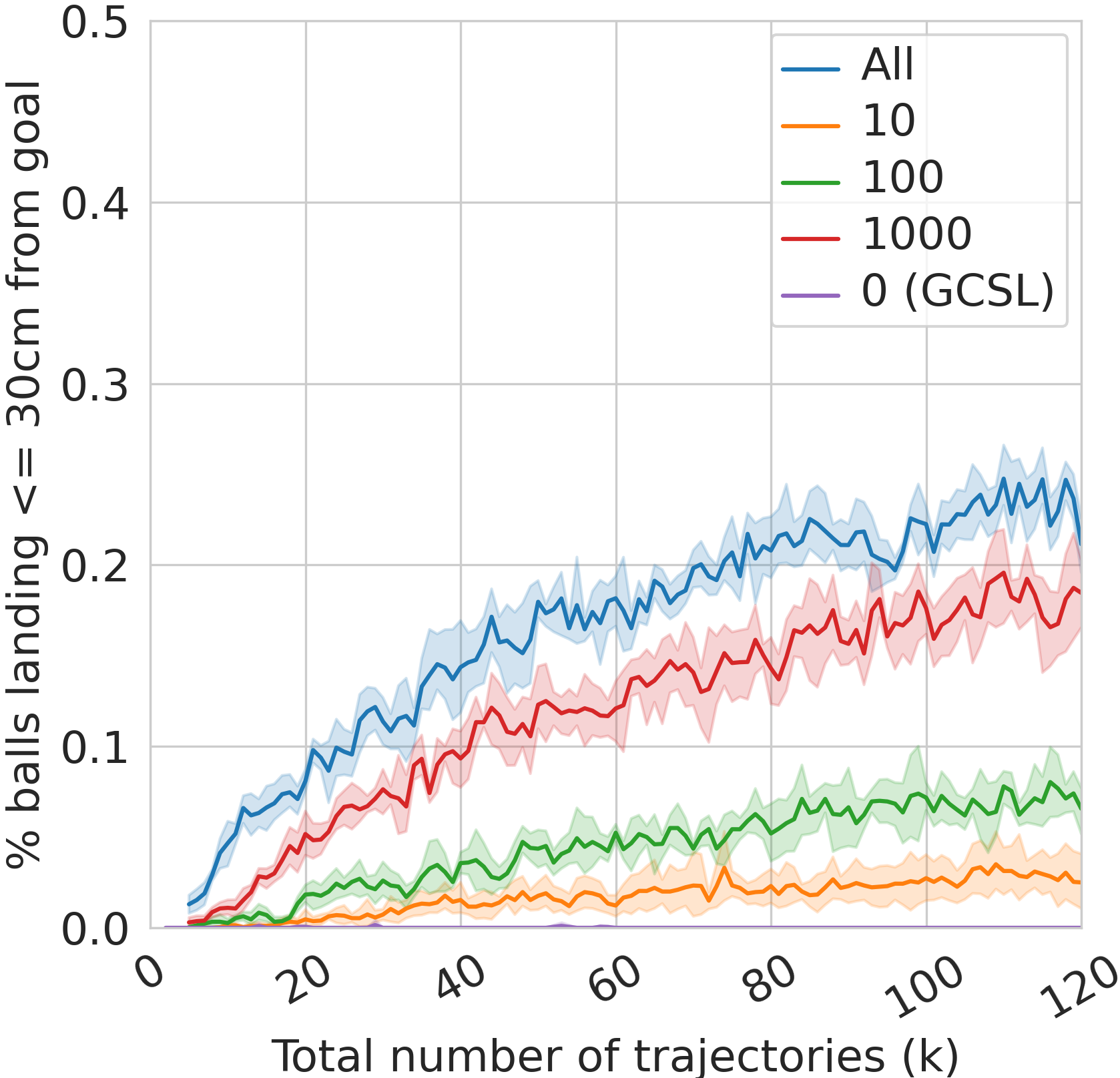}
         \caption{\% balls $\le$30cm of goal \newline  vs. total traj.}
         \label{fig:ablation_30cm}
     \end{subfigure}%
     \hfill
     \begin{subfigure}[t]{0.235\textwidth}
         \centering
         \includegraphics[width=0.8\textwidth]{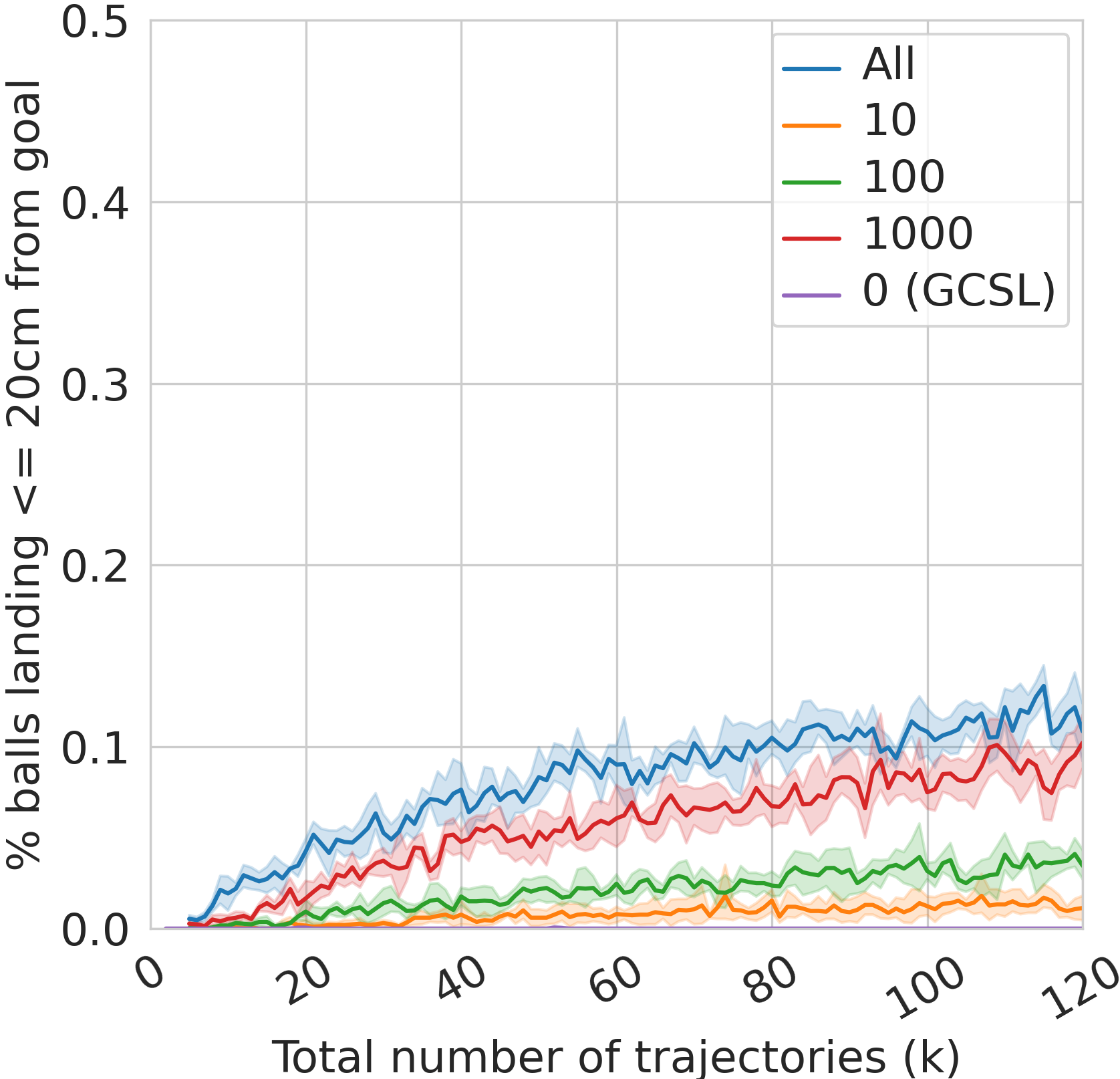}
         \caption{\% balls $\le$20cm of goal \newline vs. total traj.}
         \label{fig:ablation_20cm}
     \end{subfigure}
     \hfill
     \begin{subfigure}[t]{0.235\textwidth}
         \centering
         \includegraphics[width=0.8\textwidth]{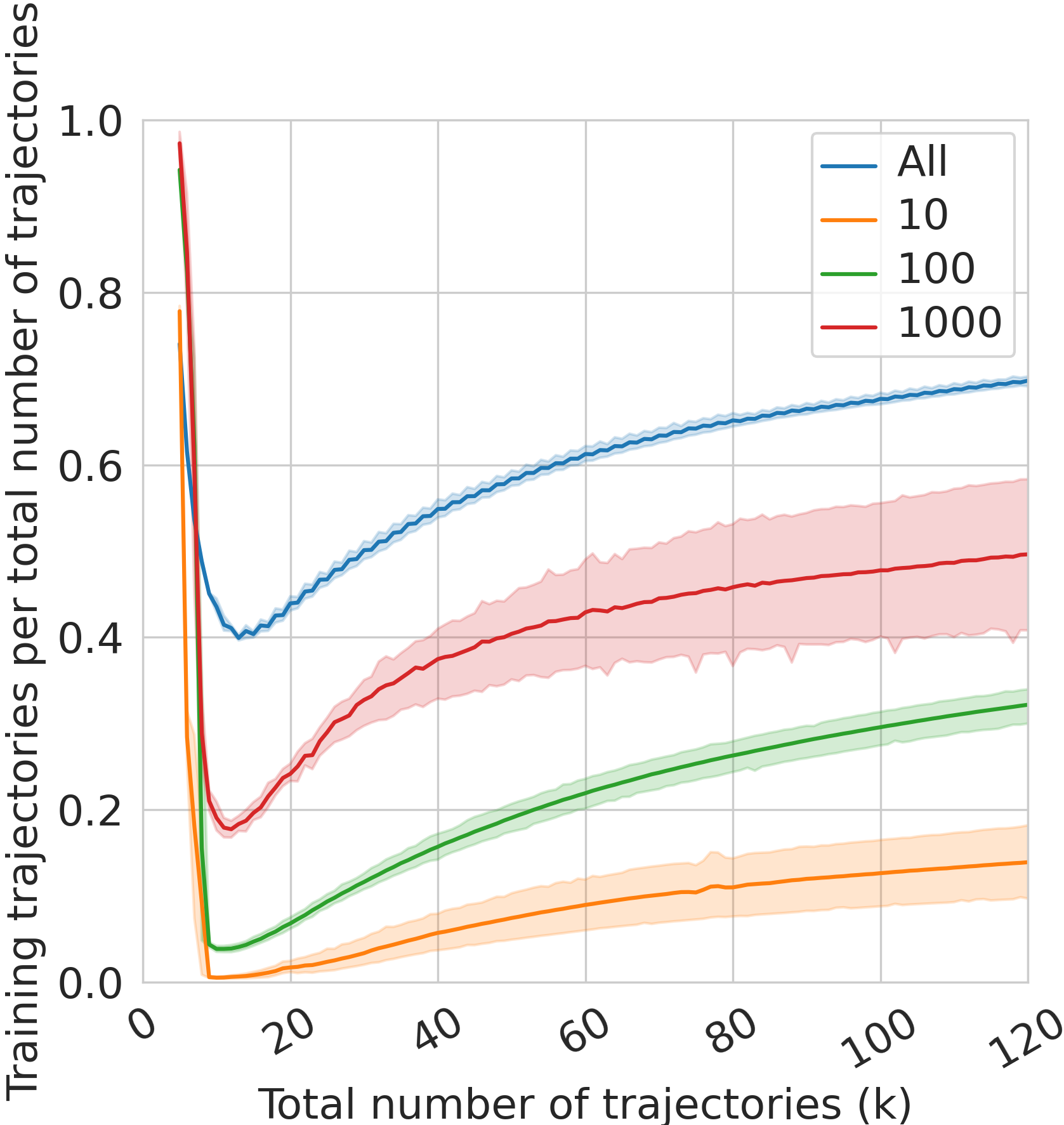}
         \caption{Training examples \newline  vs. total traj}
         \label{fig:ablation_examples_cached}
     \end{subfigure}
\caption{Simulation: Ablation study on the quantity of demonstrations: The more initial demonstrations, the more data efficient the subsequent self-supervised practice, albeit with diminishing returns.}
\label{data_efficiency}
\vspace{-4mm}
\end{figure}

\textbf{Demonstrations improve the efficiency of self-supervised practice}. We have seen that self-supervised practice improves performance over LFP in simulation. The benefits of this approach were also demonstrated by \cite{ghosh2021learning}. Here we assess the effect of demonstrations on performance by running an ablation study with 0, 10, 100, and 1000 demonstrations as shown in Figures \ref{fig:ablation_30cm} and \ref{fig:ablation_20cm}. The special case of 0 demonstrations corresponds to GCSL.

Performance improves as the number of demonstrations increases, albeit with diminishing returns. We observe that the number of ``good" trajectories that make it into the training dataset corresponds to the number of initial demonstrations. This can be seen by plotting the number of training trajectories per seen trajectories (demonstration $+$ self-practice) as shown in Figure \ref{fig:ablation_examples_cached}. The data efficiency is initially high because almost all available demonstrations are ``good", then falls because in the early stages of training because self-practice is not very effective, and gradually recovers as the policy improves. When there are fewer demonstrations the drop in efficiency is more dramatic and the differences persist throughout training. For example after 20k trajectories a policy that started with just 10 demonstrations will generate just 2 good examples for every 100 of self practice, whereas a policy that started with 1k demonstrations will generate 25. We note that this offers independent evidence for the hypothesis that demonstrations can bootstrap more efficient self-practice, explored in concurrent work \cite{dbap}.

\begin{figure}[b]
  \vspace{-6mm}
     \centering
     \begin{subfigure}[b]{0.24\textwidth}
         \centering
         \includegraphics[width=\textwidth]{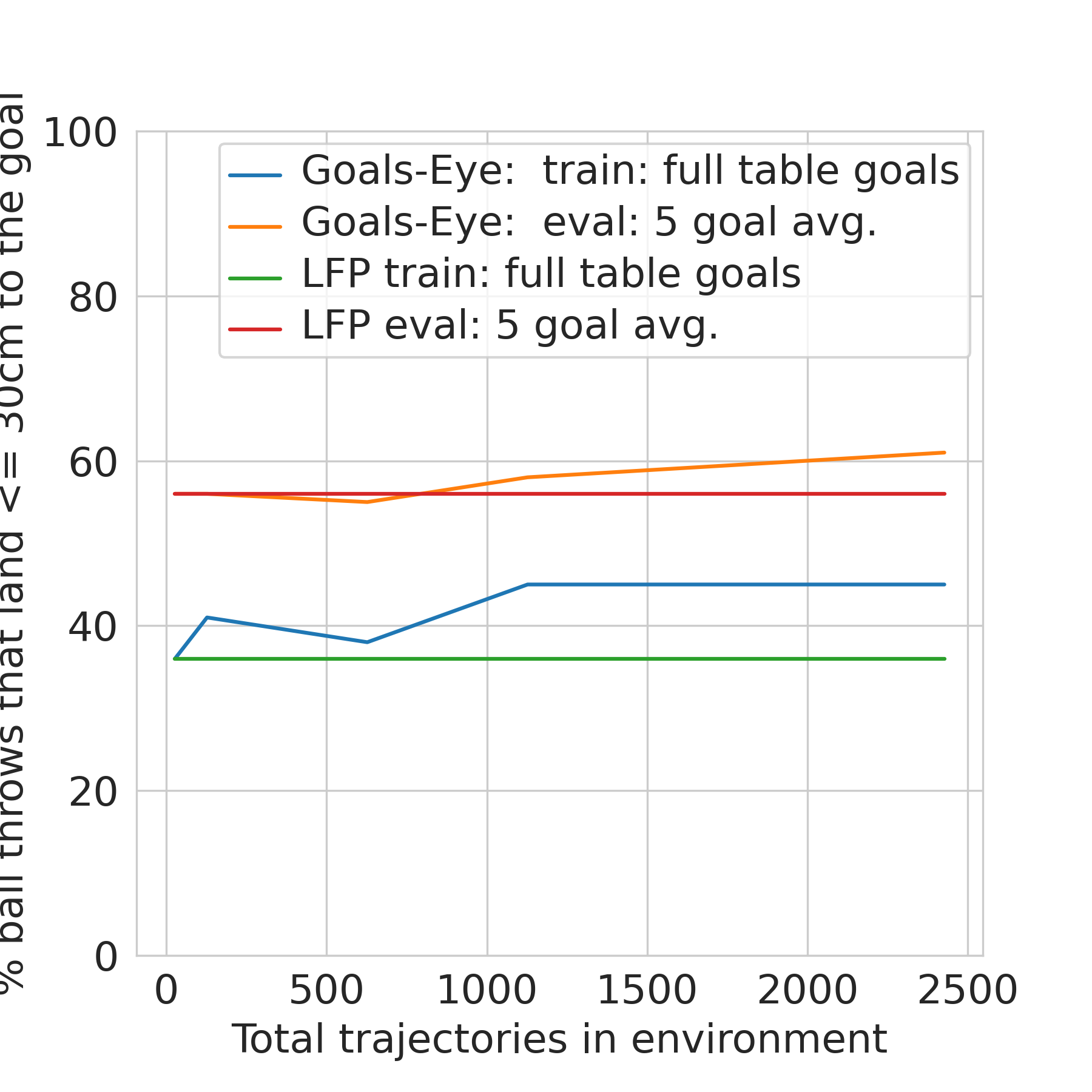}
         \caption{\% balls $\le$30cm to \newline  goal vs. total traj}
         \label{fig:real30cm}
     \end{subfigure}%
     \hfill
     \begin{subfigure}[b]{0.24\textwidth}
         \centering
         \includegraphics[width=\textwidth]{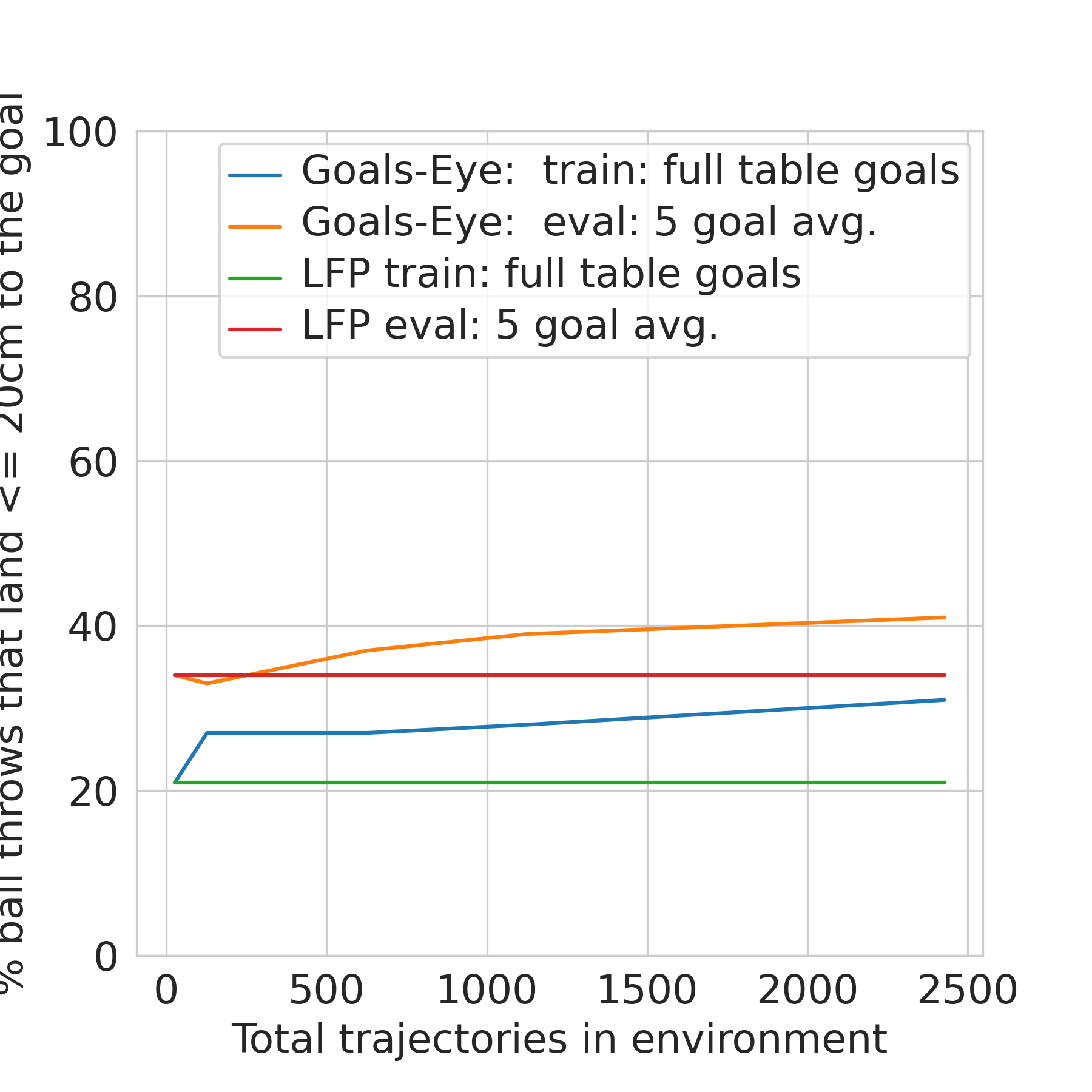}
         \caption{\% balls $\le$20cm to \newline  goal vs. total traj}
         \label{fig:real20cm}
     \end{subfigure}
     \caption{Real world: Goal reaching on the \textbf{narrow ball goal-reaching task} on a physical robot.}
     \label{main_real}
\end{figure}

\textbf{Scaling to a real-world system}. Without pre-training in simulation, we also test our approach by training a policy on the \textit{narrow ball goal-reaching task} on a physical robotic table tennis system using 27 demonstrations. We train the policy for $\approx2.5$ hours of play and $2.5$k self-practice trajectories and present our results in Figure \ref{main_real}. First we compare performance over all possible goals during training (labelled \textit{train: full table goals}). We observe that the GoalsEye policy is able to to reach 45\% of all goals to within 30cm and 31\% to within 20cm. This is a 25\% and 48\% improvement over LFP which reaches 36\% and 21\% of goals to with 30cm and 20cm respectively.

Next we set the robot the task of reaching the five specific goals (as shown in Figure \ref{fig:fivegoals}) with results labelled \textit{eval: 5 goal avg.} in Figure \ref{main_real}. Here the improvement of GoalsEye over LFP is smaller. GoalsEye reaches 61\% and 41\% of goals to within 30cm and 20cm respectively representing a 9\% and 21\% improvement over LFP which reaches 56\% and 34\% of goals to within 30cm and 20cm. \href{https://youtu.be/trUfezvmg08}{\textcolor{teal}{This}} is a real time video of the GoalsEye policy reaching each goal with a 20cm goal overlay added post-hoc.

Additionally we wished to understand how these results compared to amateur human play. To do this we set 10 humans\footnote{Note that 2 / 10 players were authors on this paper. One author-player is an advanced amateur and achieved the highest level of precision among the amateurs. The other author-player is a beginner and achieved 12\% success rate on average, comparable to the other players in their skill group. Please see the appendix for more details.} the task of reaching the same five goals. 10 humans were a random sample from a robotics lab with varying levels self-reported table tennis skill, from complete beginner to advanced amateur.

At the highest level of precision ($\le20$cm from goal) the robot outperformed all amateur humans, landing 41\% of balls on average in the target range compared to 14\% for humans on average, and 36\% for best human player (see Table \ref{tab:humancomp}). At the 30cm level of precision, the robot was $\approx2\times$ as good as humans on average, and comparable to the best human amateur. \href{https://youtu.be/9ZhtcshZTCE}{\textcolor{teal}{This}} is a video of the best amateur human reaching each goal with a 20cm goal on the table for them to aim at during evaluation.

\begin{table}[t]
\vspace{2mm}
\caption{Comparison of GoalsEye and LFP with human play. Table shows \% balls landing $\le30$cm $\vert$ $\le20$cm from goal, for 5 different goals (see Figure \ref{fig:fivegoals}). Skill was self reported. AA = amateur advanced, AI = amateur intermediate, AB = amateur beginner. A Avg. = amateur average.}
\label{tab:humancomp}
\centering
\begin{tabularx}{\columnwidth}{X|X|X|X|X|X|X}
Method                                                 & \begin{tabular}[c]{@{}l@{}}Goal\\ A\end{tabular} & \begin{tabular}[c]{@{}l@{}}Goal\\ B\end{tabular}                  & \begin{tabular}[c]{@{}l@{}}Goal\\ C\end{tabular}                  & \begin{tabular}[c]{@{}l@{}}Goal\\ D\end{tabular} & \begin{tabular}[c]{@{}l@{}}Goal\\ E\end{tabular}                  & Avg                                                               \\ \hline
LFP                                                    & 24 $\vert$ 06                                    & \textbf{74} $\vert$ 40                                                     & 64 $\vert$ \textbf{48}                                                     & 44 $\vert$ 16                                    & 76 $\vert$ 62                                                     & 56 $\vert$ 34                                                     \\ \hline
\begin{tabular}[c]{@{}l@{}}Goals\\Eye\\(ours)\end{tabular}    & 58 $\vert$ \textbf{30}                                    & \textbf{75} $\vert$ 58                                                     & 44 $\vert$ 32                                                     & 36 $\vert$14                                    & \textbf{88} $\vert$ \textbf{72}                                                     & 61 $\vert$ \textbf{41}                                                     \\ \hline
\begin{tabular}[c]{@{}l@{}}Human\\ AA\end{tabular}     & \textbf{80} $\vert$ 20                                    & 60 $\vert$ \textbf{60}                                                     & \textbf{60} $\vert$ 20                                                     & \textbf{80} $\vert$ \textbf{60}                           & 40 $\vert$ 20                                                     & \textbf{64} $\vert$ 36                                                     \\ \hline
\begin{tabular}[c]{@{}l@{}}Human\\ AI\end{tabular}     & 33 $\vert$ 05                                    & 20 $\vert$ 05                                                     & 20 $\vert$ 15                                                     & 13 $\vert$ 10                                    & 60 $\vert$ 15                                                     & 29 $\vert$ 10                                                     \\ \hline
\begin{tabular}[c]{@{}l@{}}Human\\ AB\end{tabular}     & 13 $\vert$ 08                                    & 30 $\vert$ 04                                                     & 33 $\vert$ 20                                                     & 30 $\vert$ 12                                    & 43 $\vert$ 16                                                     & 30 $\vert$ 12                                                     \\ \hline
\begin{tabular}[c]{@{}l@{}}Human\\ A. Avg\end{tabular} & 26 $\vert$ 08                                    & 30 $\vert$ 10                                                     & 32 $\vert$ 18                                                     & 30 $\vert$ 16                                    & 48 $\vert$ 16                                                     & 33 $\vert$ 14                                                     \\ \hline
\end{tabularx}
\end{table}
\section{New Results: Scaling to varied ball distributions}
\label{sec:result}
Here we outline new real world ``varied ball distribution" experiments and additional training techniques, beyond the original description of our system at publication time, that were used to obtain results in this section.

\textbf{Varied-ball goal-reaching}. Our original experiments evaluated policies on balls thrown from a fixed launch position. We present new experiments that relax this assumption, with balls now incoming from a range of launch positions and angles, where the policy must still return the ball to any location on the opponent's side of the table. This is implemented by cycling the ball thrower between three fixed locations along the x-axis. The y and z launch position of the ball remain fixed. In each of these locations the thrower is continuously rotated around the vertical axis (yaw) to vary the direction of the ball in the x axis. The thrower speed remains fixed throughout so the launch speed of the ball has a narrow range. We also initialize the robot in a more challenging initial position, a central pose with the paddle edge facing forward, requiring more dexterity to return the ball. Correspondingly we increase the control frequency of the policy for the physical robot from 20Hz to 60Hz. Finally, since this task introduces substantially more variation in the state space the bootstrap demonstration dataset for it contained 2,480 examples. A summary video of the final GoalsEye policy reaching a variety of goals can be viewed \href{https://youtu.be/SkzLjQfaQeQ}{\textcolor{teal}{here}}.

\begin{figure}[t]
     \centering
     \begin{subfigure}[t]{0.24\textwidth}
         \centering
         \includegraphics[width=\textwidth]{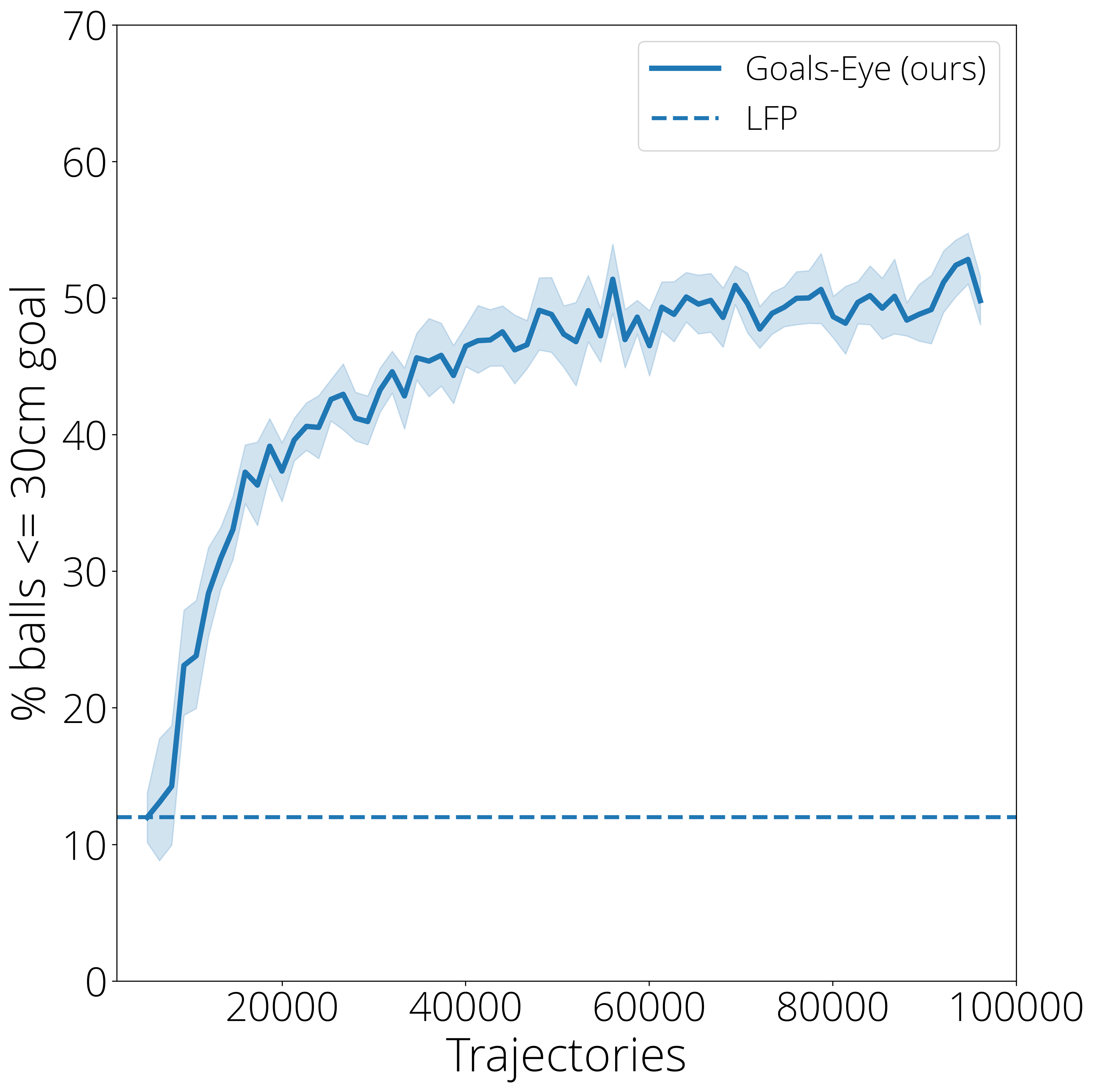}
         \caption{Simulation. N = 10 seeds.}
         \label{fig:varied_ball_sim}
     \end{subfigure}%
     \begin{subfigure}[t]{0.24\textwidth}
         \centering
         \includegraphics[width=\textwidth]{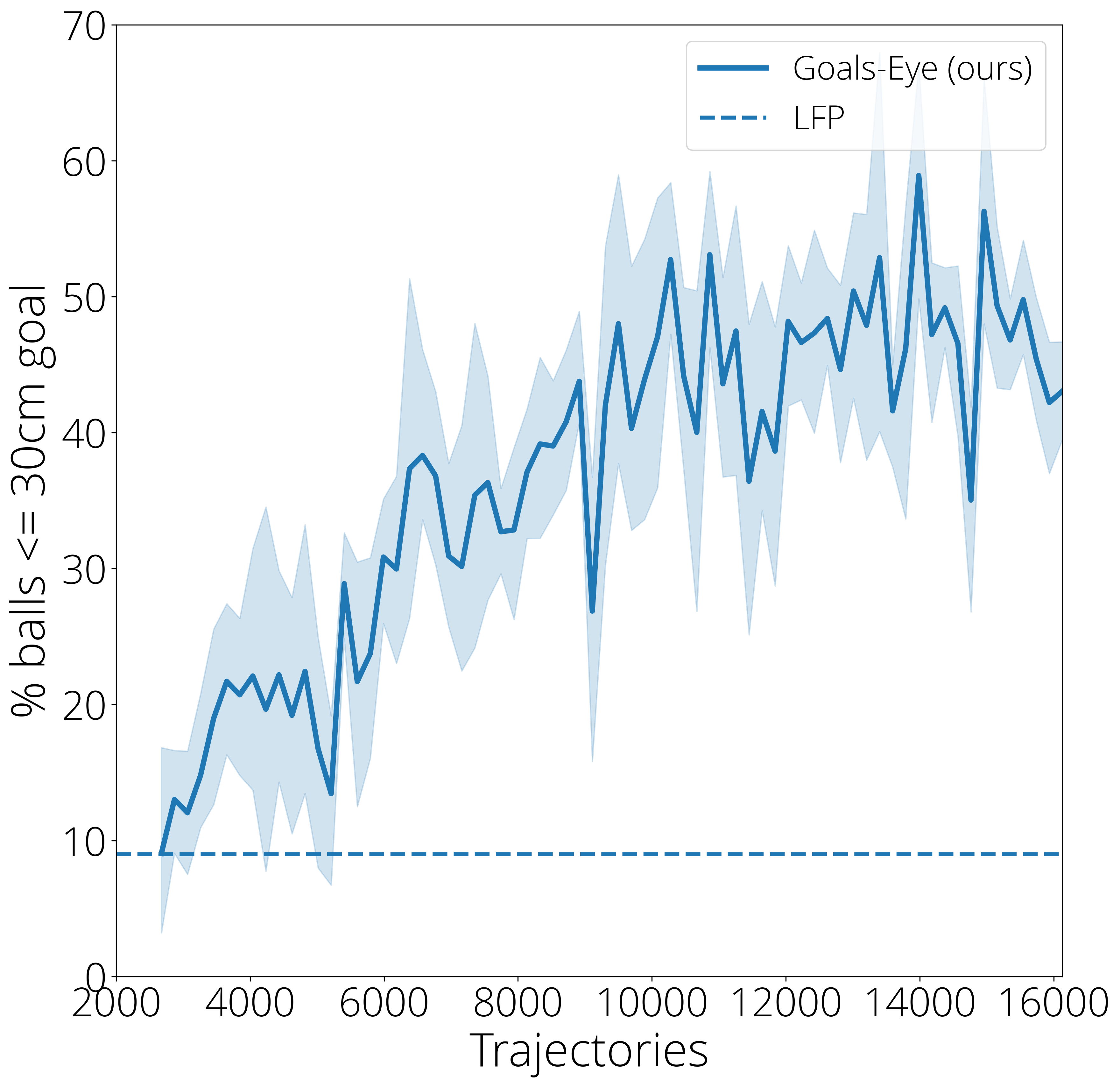}
         \caption{Physical robot. N = 5 seeds.}
         \label{fig:varied_ball_real}
     \end{subfigure}
\caption{Policies in simulation and on the physical robot reach $\approx 50$\% 30cm goal-reaching accuracy. Results are averaged over N seeds trained using data ensembling}
\label{fig:varied_ball}
\end{figure}

\textbf{Policy changes}.
We introduced three main policy learning changes when studying more varied ball throwing distributions in the real world: i) unified observation spaces between simulation and real, ii) action noise during exploration, and iii) data ensembling. We discuss all three below.

First we removed ball velocity from the simulated observation space so that a simulated policy only has access to the information available on a physical robot; robot joint positions, ball position and the goal. Then to make it easier to infer velocity and acceleration we stacked 8 time-steps to form a 2-dimensional observation of shape $(8, 13)$, where the 13 elements per time step are the robot joint positions (8), ball xyz position (3), and xy goal (2). This had an approximately neutral effect on policy performance in simulation.

We additionally experimented with adding action noise during self-supervised practice exploration, of the form $a_{\operatorname{noisy}} = a + z, z \sim \operatorname{Uniform}(-b \times a_{\operatorname{std}}, b \times a_{\operatorname{std}})$, with $b$ being a tuneable hyperparameter, and $a_{\operatorname{std}}$ being the standard deviation of actions computed over the initial training dataset. The same perturbation vector is sampled at the beginning of a rollout, and used throughout the episode unchanged. In ablations in simulation, we found this noise added during exploration lifted 30cm accuracy from 20\% to 40\%.

Finally we explored data ensembling. Here we train N models from the same pool of data (N=10 in simulation and N=5 in the real world in our case). Then each model is used for rollouts in the self-supervised practice setting. The resulting data is merged into a single dataset which is used by all N models for training. In ablations in simulation, we found that ensembling lifted 30cm accuracy from 20\% to 35\%.

The combined effect of changing the observation space, adding action noise, and data ensembling lifted 30cm accuracy by $2.5\times$ from 20\% to 50\% after 100k trajectories, motivating their use in real experiments. Figure \ref{fig:varied_ball_sim} presents the GoalsEye learning curve in simulated training. The task in simulation is the any-ball goal-reaching task, thus the results are comparable to those presented in Figure \ref{fig:sim_main30_120k}.

\begin{table}[t]
\vspace{2mm}
\caption{\textbf{Real world varied ball goal reaching}: Comparison of GoalsEye and LFP with human play on a harder task with a varied incoming ball distribution. Table shows \% balls landing $\le30$cm $\vert$ $\le20$cm from goal, for 5 different goals (see Figure \ref{fig:fivegoals}). Skill was self reported. AA = amateur advanced, AI = amateur intermediate, AB = amateur beginner. A Avg. = amateur average.}
\label{tab:humancomp_hard}
\centering
\begin{tabularx}{\columnwidth}{X|X|X|X|X|X|X}
Method & \begin{tabular}[c]{@{}l@{}}Goal\\ A\end{tabular} & \begin{tabular}[c]{@{}l@{}}Goal\\ B\end{tabular} & \begin{tabular}[c]{@{}l@{}}Goal\\ C\end{tabular} & \begin{tabular}[c]{@{}l@{}}Goal\\ D\end{tabular} & \begin{tabular}[c]{@{}l@{}}Goal\\ E\end{tabular} & Avg \\ \hline
\begin{tabular}[c]{@{}l@{}}Human\\ AA\end{tabular} & \textbf{67} $\vert$ \textbf{33} & \textbf{47} $\vert$ \textbf{27} & \textbf{100}$\vert$\textbf{87} & \textbf{80} $\vert$ \textbf{53} & \textbf{73} $\vert$ \textbf{47} & \textbf{73} $\vert$ \textbf{49} \\ \hline
\begin{tabular}[c]{@{}l@{}}Human\\ AI\end{tabular} & 23 $\vert$ 10 & 33 $\vert$ 18 & 49 $\vert$ 24 & 56 $\vert$ 34 & 42 $\vert$ 19 & 41 $\vert$ 21 \\ \hline
\begin{tabular}[c]{@{}l@{}}Human\\ AB\end{tabular} & 16 $\vert$ 11 & 13 $\vert$ 07 & 11 $\vert$ 09 & 27 $\vert$ 16 & 07 $\vert$ 02 & 15 $\vert$ 09 \\ \hline
\begin{tabular}[c]{@{}l@{}}Human\\ A. Avg\end{tabular} & 25 $\vert$ 13 & 29 $\vert$ 15 & 43 $\vert$ 26 & 49 $\vert$ 31 & 35 $\vert$ 17 & 36 $\vert$ 20 \\ \hline
LFP  & 05 $\vert$ 02 & 00 $\vert$ 00 & 24 $\vert$ 13 & 02 $\vert$ 00 & 23 $\vert$ 11 & 11 $\vert$ 05 \\ \hline
\begin{tabular}[c]{@{}l@{}}Goals\\ Eye\\ (ours)\end{tabular} & 23 $\vert$ 19 & 48 $\vert$ 33 & 32 $\vert$ 12 & 27 $\vert$18 & 68 $\vert$ 38 & 40 $\vert$ 24 \\ \hline \hline

\begin{tabular}[c]{@{}l@{}}\\ Our \\ \%-gap \\from \\human \\average \end{tabular} & -2 $\vert$ +6 & +19 $\vert$ +18 & -11 $\vert$ -14 & -22 $\vert$ -13 & +33 $\vert$ +21 & +3.4 $\vert$ +3.6 \\ \hline

\end{tabularx}
\end{table}

\textbf{Real world varied-ball results}.
We present results for real world varied-ball goal reaching in Table~\ref{tab:humancomp_hard}, comparing our policy with LFP as well as human players attempting the same task. We trained the GoalsEye policy for $\approx$16k trajectories, $\approx$15 hours of wall clock time. Policy progress during training is presented in Figure \ref{fig:varied_ball_real}. There were 2,480 initial demonstrations and the remaining 13.5k trajectories were from self-practice. On this task GoalsEye improves dramatically over LFP with 40\% of balls landing within 30cm and 24\% within 20cm on average compared with 11\% and 5\% for LFP respectively. This is unlike the narrow-ball goal reaching task in which GoalsEye yielded only moderate improvements over LFP on a physical robot. We hypothesize this is because the narrow-ball task is easier and consequently the LFP baseline was stronger, resulting in much less room for GoalsEye to improve on. Please see \href{https://youtu.be/qp3XAWORiBo}{\textcolor{teal}{this}} video for a visual comparison of the policy after training on just the initial demonstration dataset (i.e. LFP) and after additionally self-practicing (GoalsEye).

The human players have varied levels of self-reported proficiency at table tennis. We see that despite not reaching the goal-reaching performance of the advanced amateur human, GoalsEye obtains a lift of 3.4\% for balls landed within 30cm and a lift of 3.6\% for balls landed within 20cm, over average human performance. Especially under the harder varied-ball throwing format, we believe that this performance competitive with average human behavior indicates that the presented method obtains a strong base capacity for goal-directed behavior in this setting.

\section{Conclusion and Future Work}
\label{sec:conclusion}
In this paper, we studied the difficult setting of high-acceleration goal conditioned table tennis on a physical robot. We investigated a number of IL and RL techniques for goal conditioned learning, seeking the simplest possible combination that is capable of learning in this setup. Surprisingly, we found that the synthesis of two recent goal conditioned imitation learning approaches performed best, both in extensive simulated experiments and in the real world, ultimately matching or beating human amateur performance in hitting balls to specific targets. The experiments in simulation showcased the sample efficiency of this approach over RL methods, and highlighted the benefits of iterative self-supervised improvement over pure IL methods. Our experiments in the real world demonstrated for the first time, to our knowledge, that iterative imitation learning can continuously improve in the real world beyond an initial undirected bootstrap dataset, sidestepping the complexities of reinforcement learning (e.g. exploration, reward shaping, sim-to-real transfer), and excel at a dynamic tasks requiring precision.

%%%%%%%%%%%%%%%%%%%%%%%%%%%%%%%%%%%%%%%%%%%%%%%%%%%%%%%%%%%%%%%%%%%%%%%%%%%%%%%%

%%%%%%%%%%%%%%%%%%%%%%%%%%%%%%%%%%%%%%%%%%%%%%%%%%%%%%%%%%%%%%%%%%%%%%%%%%%%%%%%

%%%%%%%%%%%%%%%%%%%%%%%%%%%%%%%%%%%%%%%%%%%%%%%%%%%%%%%%%%%%%%%%%%%%%%%%%%%%%%%%
\section{Appendix}\label{sec:appendix}

\subsection{Submission video}\label{appendix:video}

In the video demo submitted with this manuscript (\href{https://youtu.be/trUfezvmg08}{\textcolor{teal}{robot}}, \href{https://youtu.be/9ZhtcshZTCE}{\textcolor{teal}{human}} evaluated on the narrow-ball task), we evaluated all the human players and the robot for the same set of 5 goals (A - E), with each goal attempted 5 times. The goal being attempted is displayed with a circle, which represents an area within 20 centimeters of the goal. \textit{Note that the circle is added virtually for the robot play whereas for the human play, there was actually a physical circle to help the humans aim correctly.} After each ball throw, the circle is virtually animated  green if the ball landed successfully within 20 centimeters of the goal, and animated red in case of a failure to do so.

Note on the harder varied ball distribution task the humans and robot players had 15 attempts per goal. See Section \ref{app:eval_details} for details.

\subsection{Algorithm Baselines}

\begin{figure}[t]
     \centering
     \begin{subfigure}[t]{0.24\textwidth}
         \centering
         \includegraphics[width=\textwidth]{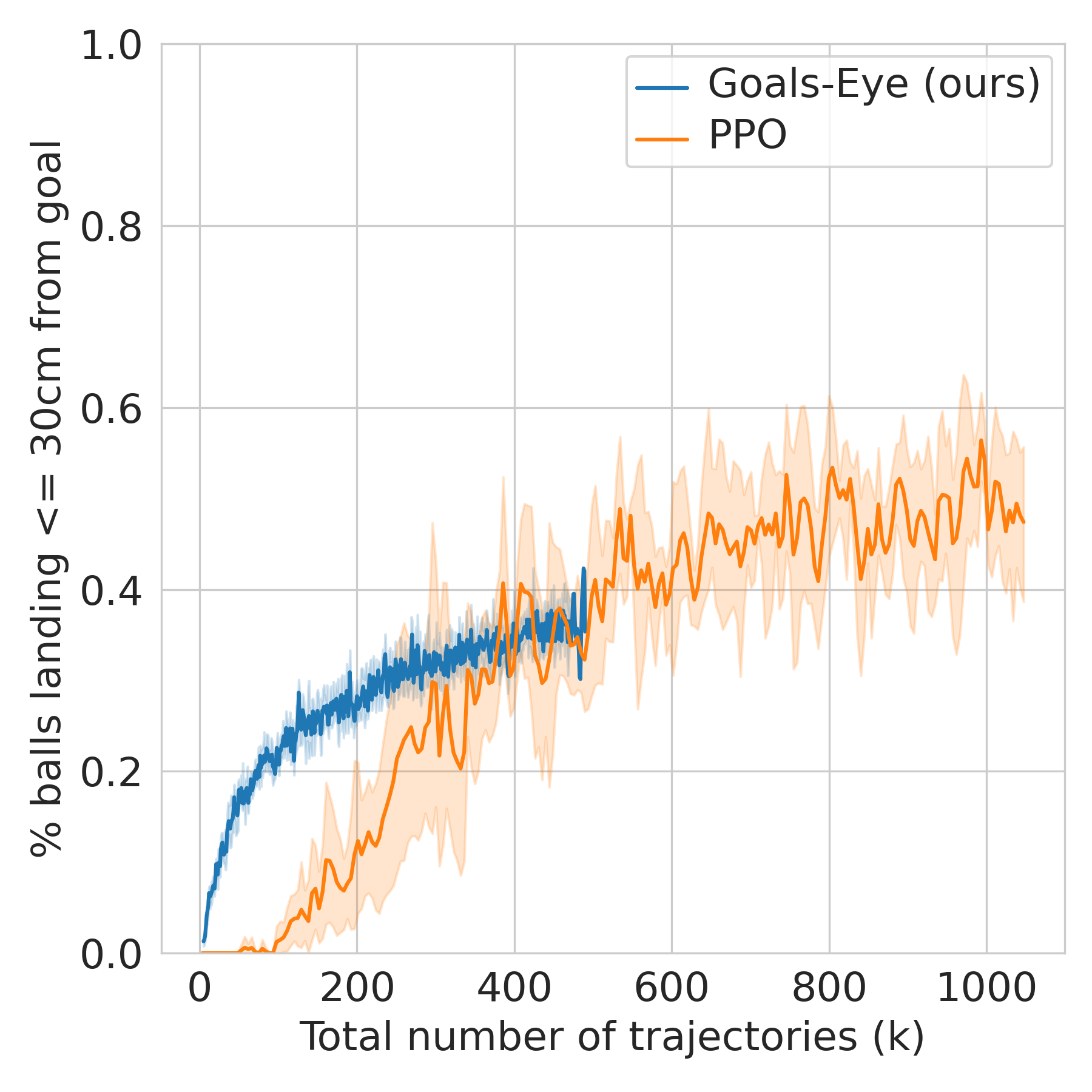}
         \caption{GoalsEye \& PPO (k)}
         \label{fig_app:sim_app_30_GCBC_PPO}
     \end{subfigure}%
     \begin{subfigure}[t]{0.24\textwidth}
         \centering
         \includegraphics[width=\textwidth]{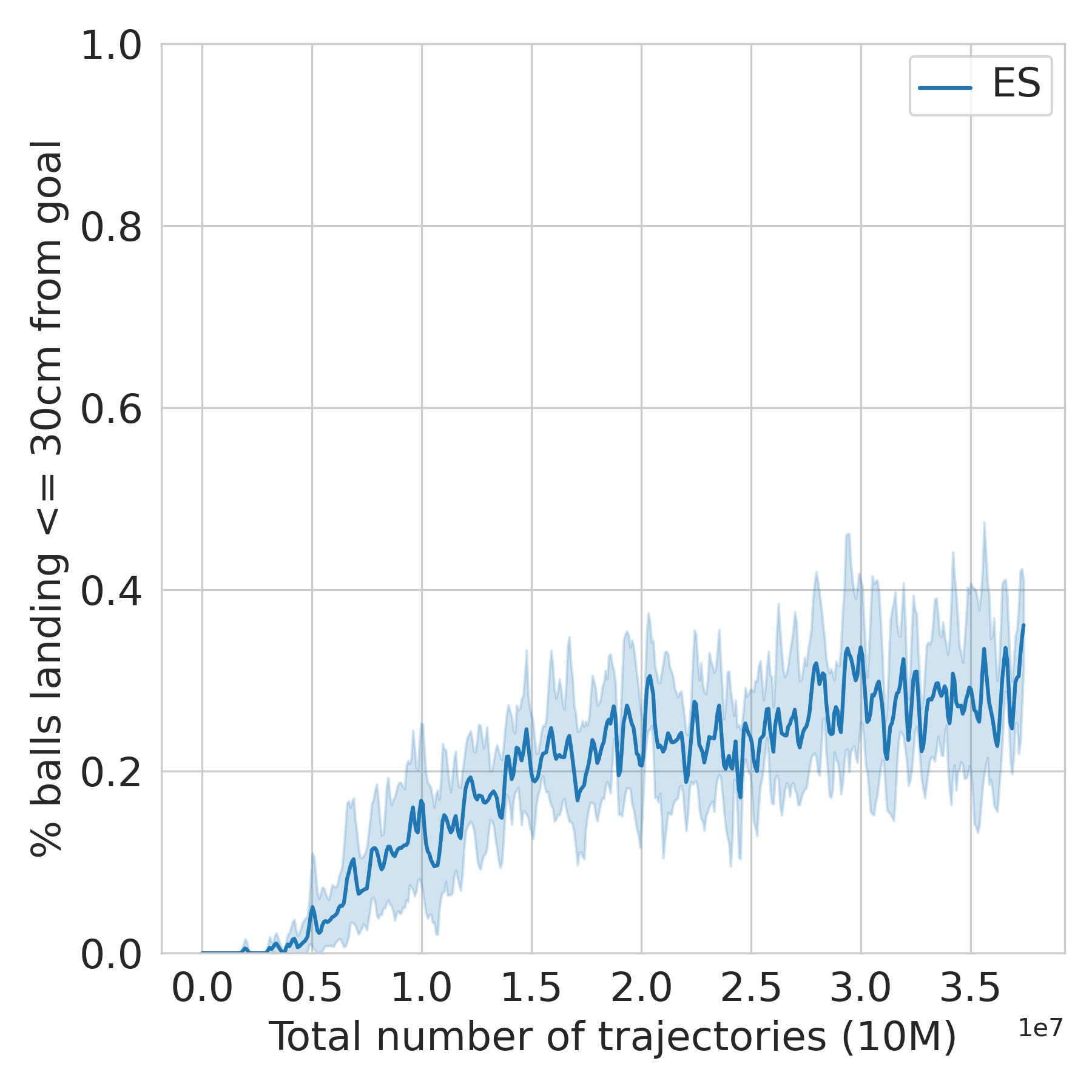}
         \caption{ES (10M)}
         \label{fig_app:sim_app_30_ES}
     \end{subfigure}
\caption{Simulation: Extended learning curves on the any-ball goal-reaching task in simulation. Results report the \% balls $\le$30cm to goal vs. total trajectories and are averaged over 5 seeds. Note the trajectory scale is in thousands of episodes for GoalsEye and PPO, and tens of millions of episodes (10M) for ES.}
\label{app:main_sim_ext}
% \vspace{-3mm}
\end{figure}

Figure \ref{fig_app:sim_app_30_GCBC_PPO} shows extended learning curves for GoalsEye and PPO which have been trained for 500k - 1M trajectories. Figure \ref{fig_app:sim_app_30_ES} shows extended learning curve for ES trained for 35M trajectories. We also continued training Qt-Opt and SAC policies for up to 700k trajectories but were not able to achieve any non zero scores.

\subsection{Human Baselines}\label{sec:human_base}

\subsubsection{Limitations of human baselines}

It it worth noting that during our human evaluation, the amateur players were not given the opportunity to warm up, which is consistent with our evaluation of the robot where no self-calibration was done before roll-outs. We hypothesize that with sufficient time to warm up, it is possible for same player to perform significantly better on the same task. Also, as noted in Section \ref{appendix:video}, for human players we had placed a physical circle on the table to help them aim better whereas for the robot play, there was nothing physical on the table. In the future, we look forward to further exploring different settings in evaluation, including creating an evaluation process allowing human players to warm up, exploring benefits from self-calibration for the robot, and designing a "fair" comparison between human and machine which includes warm-ups.

\subsubsection{Details on human data collection}

Here we include more details on the human play results presented in Table \ref{tab:humancomp}
AA = amateur advanced, AI = amateur intermediate, AB = amateur beginner. A Avg. = amateur average. For $\le30$cm, AA = 1 player, AI = 3 players, AB = 6 players. For $\le20$cm, AA = 1 player, AI = 4 players, AB = 5 players.  Human A Avg.score was averaged over 10 different amateur players who each made 5 attempts to return a ball to each goal, making 50 balls per goal in total.

\subsubsection{Details on human and robot evals: harder varied ball distribution task}
\label{app:eval_details}

Here we include more details on the human play results presented in Table \ref{tab:humancomp_hard}. For $\le30$cm and $\le20$cm, AA = 1 player, AI = 3 players, AB = 6 players. Human A Avg. score was averaged over 10 amateur players who each made 15 attempts to return a ball to each goal, with 5 attempts from each position of the ball thrower whilst it was moving around the vertical axis (yaw), making 150 balls per goal in total. Note that the amateur players were a different set of 10 players (also randomly sampled from a robotics lab) compared to the players who participated in the human evaluations on the easier narrow-ball task. This is due to the time elapsed between the first and second set of evaluations. Similarly to the previous human evaluation 2/10 players were authors on this paper.

\subsection{Author Contributions}\label{sec:author_cont}

\begin{itemize}
\item Tianli Ding wrote the training and evaluation codebase, proposed and developed the self-supervised practice framework, conducted research and ran experiments in simulation and on the real robot to reach the reported performance, coordinated human evaluation.
\item Laura Graesser ran simulation experiments, wrote the related work, method and results sections, worked on the paper narrative, edited the paper, edited videos, helped with the environment set up, advised on experiments, worked on simulator and hardware software implementation.
\item Saminda Abeyruwan ran ES baselines, ran ES and PPO evals, worked on simulator and hardware software implementation especially the gym and actor APIs, helped with real robot experiments.
\item David B. D'Ambrosio worked on simulator and hardware software implementation especially the vision system, helped with real robot experiments, wrote the hardware section of the paper.
\item Anish Shankar worked on simulator and hardware software implementation especially the overall system performance, helped tding with real robot experiments.
\item Pierre Sermanet advised on project direction and experiments, edited the video submission, edited the paper.
\item Pannag R. Sanketi managed part of the team, advised on research, helped set the research direction and wrote parts of the paper.
\item Corey Lynch advised on project direction and experiments, developed the paper story and structure, wrote the introduction, edited the paper.
\end{itemize}

\section*{ACKNOWLEDGMENT}

We thank Jonathan Tompson for their helpful feedback on the manuscript, Alex Bewley for his work on the ball perception system, and Michael Ahn, Sherry Moore, Ken Oslund, and Grace Vesom for their work on the robot control stack. We also thank thank Jon Abelian, Omar Cortes, Gus Kouretas, Thinh Nguyen, and Krista Reymann for their help maintaining our robotic system.

% %%%%%%%%%%%%%%%%%%%%%%%%%%%%%%%%%%%%%%%%%%%%%%%%%%%%%%%%%%%%%%%%%%%%%%%%%%%%%%%%

\addtolength{\textheight}{-12cm} 


\begin{thebibliography}{99}

\bibitem{rubikscube}OpenAI, Akkaya, I., Andrychowicz, M., Chociej, M., Litwin, M., McGrew, B., Petron, A., Paino, A., Plappert, M., Powell, G., Ribas, R., Schneider, J., Tezak, N., Tworek, J., Welinder, P., Weng, L., Yuan, Q., Zaremba, W. \& Zhang, L. Solving Rubik's Cube with a Robot Hand.  (2019)
\bibitem{Mahler2019LearningAR}Mahler, J., Matl, M., Satish, V., Danielczuk, M., DeRose, B., McKinley, S. \& Goldberg, K. Learning ambidextrous robot grasping policies. {\em Science Robotics}. \textbf{4} (2019)
\bibitem{Peng2020LearningAR}Peng, X., Coumans, E., Zhang, T., Lee, T., Tan, J. \& Levine, S. Learning Agile Robotic Locomotion Skills by Imitating Animals. {\em ArXiv}. \textbf{abs/2004.00784} (2020)
\bibitem{Tang2020LearningAL}Tang, Y., Tan, J. \& Harada, T. Learning Agile Locomotion via Adversarial Training. {\em 2020 IEEE/RSJ International Conference On Intelligent Robots And Systems (IROS)}. pp. 6098-6105 (2020)
\bibitem{Kalashnikov2018QTOptSD}Kalashnikov, D., Irpan, A., Pastor, P., Ibarz, J., Herzog, A., Jang, E., Quillen, D., Holly, E., Kalakrishnan, M., Vanhoucke, V. \& Levine, S. QT-Opt: Scalable Deep Reinforcement Learning for Vision-Based Robotic Manipulation. {\em ArXiv}. \textbf{abs/1806.10293} (2018)
\bibitem{Xiao2020ThinkingWM}Xiao, T., Jang, E., Kalashnikov, D., Levine, S., Ibarz, J., Hausman, K. \& Herzog, A. Thinking While Moving: Deep Reinforcement Learning with Concurrent Control. {\em ArXiv}. \textbf{abs/2004.06089} (2020)
\bibitem{hussein2017imitation}Hussein, A., Gaber, M., Elyan, E. \& Jayne, C. Imitation learning: A survey of learning methods. {\em ACM Computing Surveys (CSUR)}. \textbf{50}, 1-35 (2017)
\bibitem{Muelling2012LearningSelectGen}Muelling, K., Kober, J., Kroemer, O. \& Peters, J. Learning to select and generalize striking movements in robot table tennis. {\em The International Journal Of Robotics Research}. (2012)
\bibitem{Ibarz2021HowTT}Ibarz, J., Tan, J., Finn, C., Kalakrishnan, M., Pastor, P. \& Levine, S. How to train your robot with deep reinforcement learning: lessons we have learned. {\em The International Journal Of Robotics Research}. \textbf{40} pp. 698 - 721 (2021)
\bibitem{Ho2020RetinaGANAO}Ho, D., Rao, K., Xu, Z., Jang, E., Khansari, M. \& Bai, Y. RetinaGAN: An Object-aware Approach to Sim-to-Real Transfer. {\em ArXiv}. \textbf{abs/2011.03148} (2020)
\bibitem{lfp}Lynch, C., Khansari, M., Xiao, T., Kumar, V., Tompson, J., Levine, S. \& Sermanet, P. Learning Latent Plans from Play. {\em Conference On Robot Learning (CoRL)}. (2019), https://arxiv.org/abs/1903.01973
\bibitem{ghosh2021learning}Ghosh, D., Gupta, A., Reddy, A., Fu, J., Devin, C., Eysenbach, B. \& Levine, S. Learning to Reach Goals via Iterated Supervised Learning. {\em International Conference On Learning Representations}. (2021), https://openreview.net/forum?id=rALA0Xo6yNJ
\bibitem{haarnoja2018soft}Haarnoja, T., Zhou, A., Hartikainen, K., Tucker, G., Ha, S., Tan, J., Kumar, V., Zhu, H., Gupta, A., Abbeel, P. \& Others Soft actor-critic algorithms and applications. {\em ArXiv Preprint ArXiv:1812.05905}. (2018)
\bibitem{dbap}Gupta, A., Lynch, C., Kinman, B., Peake, G., Levine, S., \& Hausman, K. Bootstrapped Autonomous Practicing via Multi-Task Reinforcement Learning. {\em arXiv Preprint ArXiv:2203.15755} (2022).
\bibitem{DBLP:journals/corr/abs-1906-05838}Ding, Y., Florensa, C., Phielipp, M. \& Abbeel, P. Goal-conditioned Imitation Learning. {\em CoRR}. \textbf{abs/1906.05838} (2019), http://arxiv.org/abs/1906.05838
\bibitem{HER}Andrychowicz, M., Wolski, F., Ray, A., Schneider, J., Fong, R., Welinder, P., McGrew, B., Tobin, J., Abbeel, P. \& Zaremba, W. Hindsight Experience Replay. {\em Neurips}. (2017)
\bibitem{Billingsley83}Billingsley, J. Robot ping pong. {\em Practical Computing}. (1983)
\bibitem{Knight1986PingpongplayingRC}Knight, J. \& Lowery, D. Pingpong-playing robot controlled by a microcomputer. {\em Microprocessors And Microsystems - Embedded Hardware Design}. (1986)
\bibitem{Hartley87}Hartley, J. Toshiba progress towards sensory control in real time. {\em The Industrial Robot 14-1}. pp. 50-52 (1983)
\bibitem{Hashimoto1987DevelopmentOP}Hashimoto, H., Ozaki, F. \& Osuka, K. Development of Ping-Pong Robot System Using 7 Degree of Freedom Direct Drive Robots. {\em Industrial Applications Of Robotics And Machine Vision}. (1987)
\bibitem{Muelling2010Biomem}Muelling, K., Kober, J. \& Peters, J. A biomimetic approach to robot table tennis. {\em Adaptive Behavior}. (2010)
\bibitem{omron}Kyohei, A., Masamune, N. \& Satoshi, Y. The Ping Pong Robot to Return a Ball Precisely.  (2020)
\bibitem{Miyazaki2002RealizationOT}Miyazaki, F., Takeuchi, M., Matsushima, M., Kusano, T. \& Hashimoto, T. Realization of the table tennis task based on virtual targets. {\em ICRA}. (2002)
\bibitem{Miyazaki2006LearningTD}Miyazaki, F. \& Others Learning to Dynamically Manipulate: A Table Tennis Robot Controls a Ball and Rallies with a Human Being. {\em Advances In Robot Control}. (2006)
\bibitem{Anderson1988ARP}Anderson, R. A Robot Ping-Pong Player: Experiments in Real-Time Intelligent Control. (MIT Press,1988)
\bibitem{Muelling2010SimulatingHT}Muelling, K. \& Others Simulating Human Table Tennis with a Biomimetic Robot Setup. {\em Simulation Of Adaptive Behavior}. (2010)
\bibitem{Zhu2018TowardsHL}Zhu, Y., Zhao, Y., Jin, L., Wu, J. \& Xiong, R. Towards High Level Skill Learning: Learn to Return Table Tennis Ball Using Monte-Carlo Based Policy Gradient Method. {\em IEEE International Conference On Real-time Computing And Robotics}. (2018)
\bibitem{Huang2015LearningOS}Huang, Y., Schölkopf, B. \& Peters, J. Learning optimal striking points for a ping-pong playing robot. {\em IROS}. (2015)
\bibitem{Sun2011BalanceMG}Sun, Y., Xiong, R., Zhu, Q., Wu, J. \& Chu, J. Balance motion generation for a humanoid robot playing table tennis. {\em IEEE-RAS Humanoids}. (2011)
\bibitem{Mahjourian2018HierarchicalPD}Mahjourian, R., Jaitly, N., Lazic, N., Levine, S. \& Miikkulainen, R. Hierarchical Policy Design for Sample-Efficient Learning of Robot Table Tennis Through Self-Play. {\em ArXiv:1811.12927}. (2018)
\bibitem{Matsushima2003LearningTT}Matsushima, M., Hashimoto, T. \& Miyazaki, F. Learning to the robot table tennis task-ball control and rally with a human. {\em IEEE International Conference On Systems, Man And Cybernetics}. (2003)
\bibitem{Matsushima2005ALA}Matsushima, M., Hashimoto, T., Takeuchi, M. \& Miyazaki, F. A learning approach to robotic table tennis. {\em IEEE Transactions On Robotics}. (2005)
\bibitem{Muelling2010LearningTTMOMP}Muelling, K., Kober, J. \& Peters, J. Learning table tennis with a Mixture of Motor Primitives. {\em IEEE-RAS Humanoids}. (2010)
\bibitem{Huang2016JointlyLT}Huang, Y., Buchler, D., Koç, O., Schölkopf, B. \& Peters, J. Jointly learning trajectory generation and hitting point prediction in robot table tennis. {\em IEEE-RAS Humanoids}. (2016)
\bibitem{Ko2018OnlineOT}Koç, O., Maeda, G. \& Peters, J. Online optimal trajectory generation for robot table tennis. {\em Robotics \& Autonomous Systems}. (2018)
\bibitem{Tebbe2018ATT}Tebbe, J., Gao, Y., Sastre-Rienietz, M. \& Zell, A. A Table Tennis Robot System Using an Industrial KUKA Robot Arm. {\em GCPR}. (2018)
\bibitem{Gao2019MarkerlessRP}Gao, Y., Tebbe, J., Krismer, J. \& Zell, A. Markerless Racket Pose Detection and Stroke Classification Based on Stereo Vision for Table Tennis Robots. {\em IEEE Robotic Computing}. (2019)
\bibitem{Akrour2016ModelFreeTP}Akrour, R., Abdolmaleki, A., Abdulsamad, H., Peters, J. \& Neumann, G. Model-Free Trajectory-based Policy Optimization with Monotonic Improvement. {\em J. Mach. Learn. Res.}. (2016)
\bibitem{LFSD-GT}Chen, L., Paleja, R. \& Gombolay, M. Learning from Suboptimal Demonstration via Self-Supervised Reward Regression. {\em CoRL}. (2020)
\bibitem{GaoPPOES2020}Gao, W., Graesser, L., Choromanski, K., Song, X., Lazic, N., Sanketi, P., Sindhwani, V. \& Jaitly, N. Robotic Table Tennis with Model-Free Reinforcement Learning. {\em IROS}. (2020)
\bibitem{buchler}Büchler, D., Guist, S., Calandra, R., Berenz, V., Schölkopf, B. \& Peters, J. Learning to Play Table Tennis From Scratch using Muscular Robots.  (2020)
\bibitem{SERL_tebbe}Tebbe, J., Krauch, L., Gao, Y. \& Zell, A. Sample-efficient Reinforcement Learning in Robotic Table Tennis. {\em ICRA}. (2021)
\bibitem{LHPHETSW2016ICLR}Lillicrap, T., Hunt, J., Pritzel, A., Heess, N., Erez, T., Tassa, Y., Silver, D. \& Wierstra, D. Continuous control with deep reinforcement learning. {\em ICLR}. (2016)
\bibitem{SWDRK2017arxiv}Schulman, J., Wolski, F., Dhariwal, P., Radford, A. \& Klimov, O. Proximal policy optimization algorithms. {\em ArXiv:1707.06347}. (2017)
\bibitem{WSPS2008EC}Wierstra, D., Schaul, T., Peters, J. \& Schmidhuber, J. Natural evolution strategies. {\em 2008 IEEE Congress On Evolutionary Computation}. (2008)
\bibitem{NS2017FOCM}Nesterov, Y. \& Spokoiny, V. Random Gradient-Free Minimization of Convex Functions. {\em FoCM}. (2017)
\bibitem{actionable_models}Chebotar, Y., Hausman, K., Lu, Y., Xiao, T., Kalashnikov, D., Varley, J., Irpan, A., Eysenbach, B., Julian, R., Finn, C. \& Levine, S. Actionable Models: Unsupervised Offline Reinforcement Learning of Robotic Skills.  (2021)
\bibitem{awac}Nair, A., Dalal, M., Gupta, A. \& Levine, S. Accelerating Online Reinforcement Learning with Offline Datasets.  (2020)
\bibitem{DBLP:journals/corr/abs-1807-04742}Nair, A., Pong, V., Dalal, M., Bahl, S., Lin, S. \& Levine, S. Visual Reinforcement Learning with Imagined Goals. {\em CoRR}. \textbf{abs/1807.04742} (2018), http://arxiv.org/abs/1807.04742
\bibitem{openai2021asymmetric}OpenAI, O., Plappert, M., Sampedro, R., Xu, T., Akkaya, I., Kosaraju, V., Welinder, P., D'Sa, R., Petron, A., Oliveira Pinto, H., Paino, A., Noh, H., Weng, L., Yuan, Q., Chu, C. \& Zaremba, W. Asymmetric self-play for automatic goal discovery in robotic manipulation.  (2021), https://openreview.net/forum?id=hu2aMLzOxC
\bibitem{abb2020egm}ABB Application manual - Externally Guided Motion. (Vasteras,2020)
\bibitem{coumans2019}Coumans, E. \& Bai, Y. PyBullet, a Python module for physics simulation for games, robotics and machine learning. (http://pybullet.org)
\bibitem{ChoromanskiRSTW18}Choromanski, K., Rowland, M., Sindhwani, V., Turner, R. \& Weller, A. Structured Evolution with Compact Architectures for Scalable Policy Optimization. {\em Proceedings Of The 35th International Conference On Machine Learning, ICML 2018, Stockholmsmässan, Stockholm, Sweden, July 10-15, 2018}. \textbf{80} pp. 969-977 (2018), http://proceedings.mlr.press/v80/choromanski18a.html
\bibitem{DBLP:journals/corr/SchulmanWDRK17}Schulman, J., Wolski, F., Dhariwal, P., Radford, A. \& Klimov, O. Proximal Policy Optimization Algorithms. {\em CoRR}. \textbf{abs/1707.06347} (2017), http://arxiv.org/abs/1707.06347
\bibitem{sac}Haarnoja, T., Zhou, A., Abbeel, P. \& Levine, S. Soft Actor-Critic: Off-Policy Maximum Entropy Deep Reinforcement Learning with a Stochastic Actor. {\em ICML}. (2018)
\bibitem{nair2018visual}Nair, A., Pong, V., Dalal, M., Bahl, S., Lin, S. \& Levine, S. Visual reinforcement learning with imagined goals. {\em ArXiv Preprint ArXiv:1807.04742}. (2018)
\bibitem{zhu2020ingredients}Zhu, H., Yu, J., Gupta, A., Shah, D., Hartikainen, K., Singh, A., Kumar, V. \& Levine, S. The ingredients of real-world robotic reinforcement learning. {\em ArXiv Preprint ArXiv:2004.12570}. (2020)
\bibitem{fujimoto2019benchmarking}Fujimoto, S., Conti, E., Ghavamzadeh, M. \& Pineau, J. Benchmarking batch deep reinforcement learning algorithms. {\em ArXiv Preprint ArXiv:1910.01708}. (2019)
\bibitem{dulac2020empirical}Dulac-Arnold, G., Levine, N., Mankowitz, D., Li, J., Paduraru, C., Gowal, S. \& Hester, T. An empirical investigation of the challenges of real-world reinforcement learning. {\em ArXiv Preprint ArXiv:2003.11881}. (2020)
\bibitem{agarwal2019striving}Agarwal, R., Schuurmans, D. \& Norouzi, M. Striving for simplicity in off-policy deep reinforcement learning.  (2019)
\bibitem{andrychowicz2020matters}Andrychowicz, M., Raichuk, A., Stańczyk, P., Orsini, M., Girgin, S., Marinier, R., Hussenot, L., Geist, M., Pietquin, O., Michalski, M. \& Others. What matters in on-policy reinforcement learning? a large-scale empirical study. {\em ArXiv Preprint ArXiv:2006.05990}. (2020)
\bibitem{yu2020meta}Yu, T., Quillen, D., He, Z., Julian, R., Hausman, K., Finn, C. \& Levine, S. Meta-world: A benchmark and evaluation for multi-task and meta reinforcement learning. {\em Conference On Robot Learning}. pp. 1094-1100 (2020)
\bibitem{mandlekar2021matters}Mandlekar, A., Xu, D., Wong, J., Nasiriany, S., Wang, C., Kulkarni, R., Fei-Fei, L., Savarese, S., Zhu, Y. \& Martin-Martin, R. What Matters in Learning from Offline Human Demonstrations for Robot Manipulation. {\em Conference On Robot Learning}. (2021)
\bibitem{zhang2018deep}Zhang, T., McCarthy, Z., Jow, O., Lee, D., Chen, X., Goldberg, K. \& Abbeel, P. Deep imitation learning for complex manipulation tasks from virtual reality teleoperation. {\em 2018 IEEE International Conference On Robotics And Automation (ICRA)}. pp. 5628-5635 (2018)
\bibitem{NES}Wierstra, D., Schaul, T., Peters, J. \& Schmidhuber, J. Natural Evolution Strategies. {\em 2008 IEEE Congress On Evolutionary Computation (IEEE World Congress On Computational Intelligence)}. pp. 3381-3387 (2008)
\bibitem{Nesterov2017RandomGM}Nesterov, Y. \& Spokoiny, V. Random Gradient-Free Minimization of Convex Functions. {\em Foundations Of Computational Mathematics}. \textbf{17} pp. 527-566 (2017)
\bibitem{salimans2017evolution}Salimans, T., Ho, J., Chen, X., Sidor, S. \& Sutskever, I. Evolution Strategies as a Scalable Alternative to Reinforcement Learning.  (2017)
\bibitem{ARS}Mania, H., Guy, A. \& Recht, B. Simple random search provides a competitive approach to reinforcement learning. {\em CoRR}. \textbf{abs/1803.07055} (2018), http://arxiv.org/abs/1803.07055
\bibitem{PolicySearchHRL}End, F., Akrour, R., Peters, J. \& Neumann, G. Layered direct policy search for learning hierarchical skills. {\em 2017 IEEE International Conference On Robotics And Automation (ICRA)}. pp. 6442-6448 (2017)



\end{thebibliography}
\end{document}